\documentclass{article}

\usepackage{microtype}
\usepackage{graphicx}
\usepackage{booktabs} \usepackage{wrapfig}
\usepackage{multirow}
\usepackage[T1]{fontenc}    \usepackage{subcaption}

\usepackage{hyperref}
\usepackage{xcolor}

\usepackage[accepted]{icml2024}

\usepackage{amsmath}
\usepackage{amssymb}
\usepackage{mathtools}
\usepackage{amsthm}

\usepackage[capitalize,noabbrev]{cleveref}

\usepackage{amsmath,amsfonts,bm}

\def\eqref#1{equation~\ref{#1}}

\def\1{\bm{1}}

\DeclareMathAlphabet{\mathsfit}{\encodingdefault}{\sfdefault}{m}{sl}
\SetMathAlphabet{\mathsfit}{bold}{\encodingdefault}{\sfdefault}{bx}{n}

\newcommand{\E}{\mathbb{E}}

\theoremstyle{plain}

\theoremstyle{definition}

\theoremstyle{remark}

\usepackage[textsize=tiny]{todonotes}

\icmltitlerunning{Fast Peer Adaptation with Context-aware Exploration}

\begin{document}

\twocolumn[
\icmltitle{Fast Peer Adaptation with Context-aware Exploration}

\icmlsetsymbol{equal}{*}

\begin{icmlauthorlist}
\icmlauthor{Long Ma*}{aais,bigai}
\icmlauthor{Yuanfei Wang*}{cfcs,bigai}
\icmlauthor{Fangwei Zhong}{ist,bigai}
\icmlauthor{Song-Chun Zhu}{iai,ist,bigai}
\icmlauthor{Yizhou Wang}{cfcs,iai,cvt,bigai} \end{icmlauthorlist}

\icmlaffiliation{aais}{Academy for Advanced Interdisciplinary Studies, Peking University}
\icmlaffiliation{ist}{School of Intelligence Science and Technology, Peking University}
\icmlaffiliation{cfcs}{Center on Frontiers of Computing Studies, School of Computer Science, Peking University}
\icmlaffiliation{iai}{Inst. for Artificial Intelligence, Peking University}
\icmlaffiliation{cvt}{Nat'l Eng. Research Center of Visual Technology, Peking University}
\icmlaffiliation{bigai}{Nat'l Key Laboratory of General Artificial Intelligence, BIGAI\&PKU}

\icmlcorrespondingauthor{Fangwei Zhong}{zfw@pku.edu.cn}

\icmlkeywords{Reinforcement learning}

\vskip 0.3in
]

\printAffiliationsAndNotice{\icmlEqualContribution} 
\begin{abstract}
Fast adapting to unknown peers (partners or opponents) with different strategies is a key challenge in multi-agent games.
To do so, it is crucial for the agent to probe and identify the peer’s strategy efficiently, as this is the prerequisite for carrying out the best response in adaptation.
However, exploring the strategies of unknown peers is difficult, especially when the games are partially observable and have a long horizon.
In this paper, we propose a peer identification reward, which rewards the learning agent based on how well it can identify the behavior pattern of the peer over the historical context, such as the observation over multiple episodes.
This reward motivates the agent to learn a context-aware policy for effective exploration and fast adaptation, i.e., to actively seek and collect informative feedback from peers when uncertain about their policies and to exploit the context to perform the best response when confident.
We evaluate our method on diverse testbeds that involve competitive (Kuhn Poker), cooperative (PO-Overcooked), or mixed (Predator-Prey-W) games with peer agents.
We demonstrate that our method induces more active exploration behavior, achieving faster adaptation and better outcomes than existing methods \footnote{Project page: \url{https://sites.google.com/view/peer-adaptation}}.

\end{abstract}

\section{Introduction}

\begin{figure}[t]
    \centering
    \includegraphics[width=\linewidth]{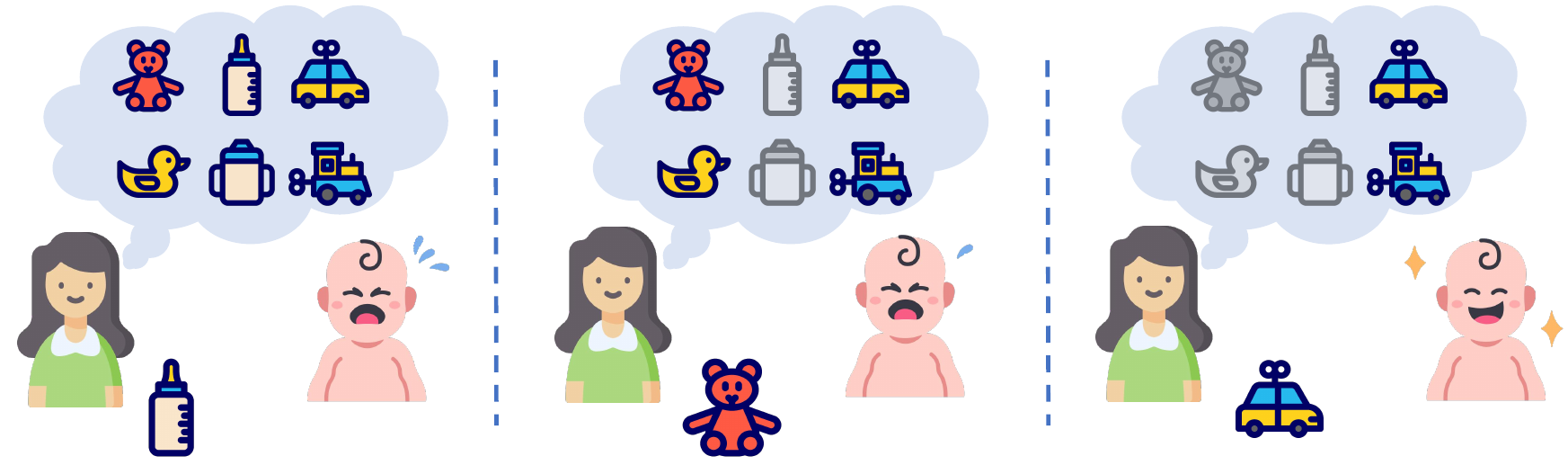}
    \vspace{-0.6cm}
    \caption{An example of fast peer adaptation with experiences from online interaction, where a mother employs her prior experiences with her baby as contextual cues to determine the appropriate item to offer and further explore the baby. In the initial encounter, having observed the baby's disinterest in the milk bottle, the mother infers that the baby is not hungry and suggests a toy as an alternative. Despite the initial unfavorable response to the teddy bear, there is a discernible improvement in the baby's reaction, ultimately leading the mother to successfully choose a toy car in their third interaction.}
    \label{fig:demo}
    \vspace{-0.4cm}
\end{figure}

\emph{Fast adaptation to diverse peers} is a key ability for social agents, who often face unknown peers with different cooperative or competitive strategies in multi-agent games.
Thus, the agents’ performance hinges on \emph{how quickly and effectively they can adapt to these peers}.
This requires the agents to efficiently probe and identify the peer’s strategy and respond with the optimal strategies accordingly. 
This is essential for achieving successful cooperation or exploitation in various domains.
For example, in board and card games~\citep{doi:10.1126/science.aay2400, silver2017mastering, hu2020other}, agents need to adjust to the skill and style of their opponents or teammates, such as bluffing, coordination, or signaling. In DOTA~\citep{berner2019dota}, agents need to cope with the dynamic strategies and tactics of the enemy team, such as counter-picking, ganking, or pushing.

\begin{figure*}[!t]
    \centering
        \includegraphics[width=\textwidth]{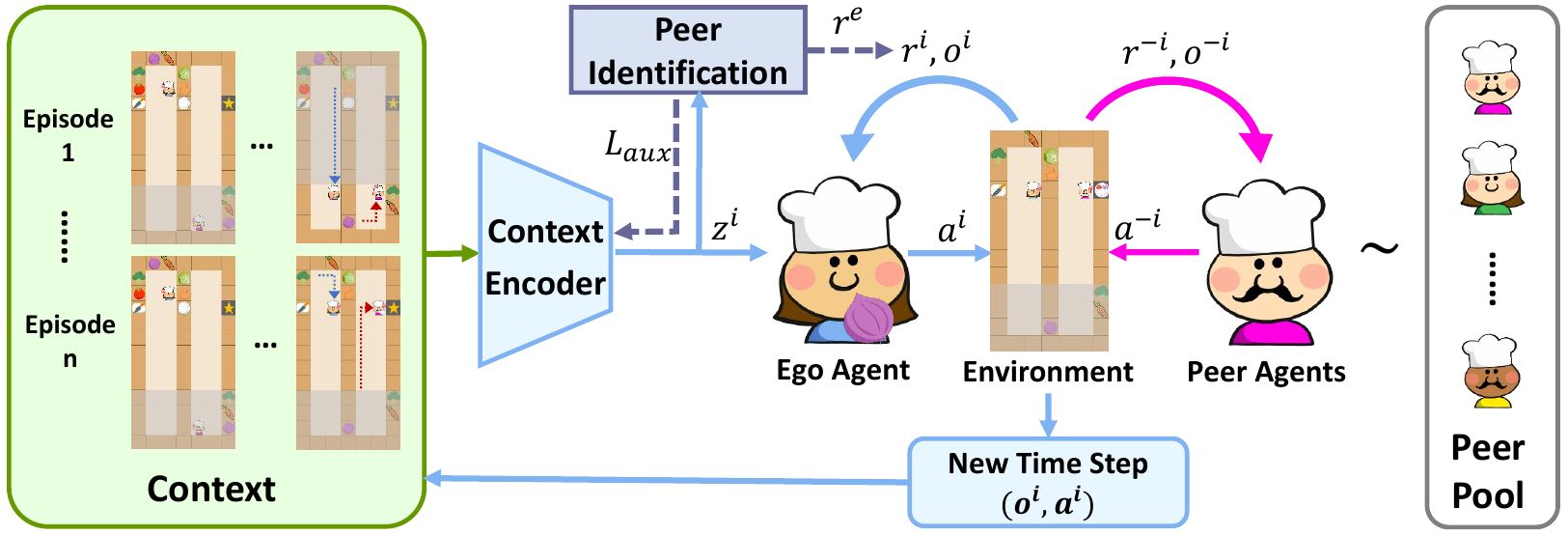}
    \vspace{-0.6cm}
    \caption{
    Illustration of PACE.
    The ego agent (left) is trained against a diverse pool of peers (right) during training.
    Conditioned on the past episodes, the ego agent proposes new actions to explore the peer or exploit the best response.
    The peer identification objective backpropagates to the context encoder and generates exploration reward for the policy to maximize mutual information.
                    }
        \label{fig:arch}
    \vspace{-0.3cm}
\end{figure*}

Previous works overlooked the role of efficient exploration in obtaining informative feedback for fast peer adaptation. They mainly focused on opponent modeling~\cite{he2016opponent, raileanu2018modeling, wangtom2c, papoudakis2021agent, yu2022model, albrecht2018autonomous}, assuming that the ego agents can readily observe the others’ behaviors and infer the best response accordingly. However, this assumption may not hold in partially observable environments, which are more realistic and challenging for multi-agent games. Therefore, the ego agent needs to learn an exploratory policy that can actively induce the game states that reveal the peers’ preferences and strategies.

Learning to explore is challenging in peer adaptation, as it involves trading off short-term and long-term outcomes under a limited context and without explicit reward signals. Figure.~\ref{fig:demo} shows an example of exploring the baby’s preference when soothing a crying baby. 
Likewise, in card games, players may call bluffs to learn opponents’ tendencies, risking short-term losses but enhancing long-term strategy.
Similarly, in DOTA, human players use exploration behaviors such as warding to obtain enemy states, sacrificing time and resources but enabling counter-strategy.

Motivated by this, we propose a fast \emph{\textbf{P}eer \textbf{A}daptation method with \textbf{C}ontext-aware \textbf{E}xploration} \textbf{(PACE)} that encourages context-aware exploration of peer agents while optimizing total returns over multiple episodes of interaction during adaptation.
We introduce \textbf{peer identification} as an auxiliary task that facilitates the recognition of the peers' strategies based on the observed context.
During training, a peer identifier is trained to estimate a representation of the current training peer policy given the current interaction context.
With the identification, a peer identification reward is used to guide the learning of the ego policy, promoting exploratory behaviors and the emergence of informative contexts about peer policies.
The ego policy is context-aware and trained to optimize a multi-episode return objective, striking an overall exploration-exploitation balance across multiple episodes of online adaptation.
The context encoder is shared between the ego policy and the auxiliary task to provide better peer representations for policy learning.

We conduct experiments in a competitive environment (Kuhn Poker), a cooperative environment (PO-Overcooked), and a mixed environment with both cooperative and competitive peers (Predator-Prey-W).
We show that PACE adapts faster and achieves higher returns than existing methods when facing unknown opponents or collaborators.
In further ablation studies, we analyze the effects of the proposed auxiliary task and the corresponding intrinsic reward.
A t-SNE visualization of the latent embedding shows that the PACE agent quickly distinguishes between the peer policies.

In summary, our main contributions are three-fold.
1) We investigate the peer adaptation problem in detail and propose the peer identification reward to address the insufficient exploration problem.
2) We introduce a practical context-aware policy learning method that optimizes cross-episode task rewards with exploration rewards against a diverse pool of peers, promoting exploration-exploitation balance.
3) We empirically validate our method in a competitive card game (Kuhn Poker), a cooperative game (PO-Overcooked), and a mixed environment (Predator-Prey-W), showing that our context-aware policy can quickly adapt to unknown peers and achieve high performance.

\vspace{-0.2cm}
\section{Related Work}
\vspace{-0.2cm}

\textbf{Fast Adaptation.} Fast Adaptation includes adaptation to new environments (tasks)~\citep{finn2017model, raileanu2020fast, zuo2019craves, laskin2022context, luo2022adapt} and new agents~\citep{stone2010ad, ravula2019ad,rakelly2019efficient, zhu2021few, zhong2019ad, zhong2021towards, rahman2021towards, yan2023efficient}. In this paper, we consider the problem of fast adaptation to unknown agents. 
Meta-learning methods~\citep{al2018continuous, kim2021policy} compute meta-multiagent policy gradient during interaction to adapt the policy accordingly. Bayesian inference~\citep{zintgraf2021deep} has been investigated for updating the belief about other agents to effectively respond to them.
Some methods~\citep{zhang2023fast, gu2021online} utilize Value Decomposition with teammate context modeling to achieve online adaptation. Modeling of other agents can help improve the adaptation to them~\citep{he2016opponent, raileanu2018modeling, wangtom2c, papoudakis2021agent, yu2022model, albrecht2018autonomous, fu2022greedy, xie2021learning}. However, the previous methods assume that the contexts about peers are easy to obtain, while we focus on partially observable games that require active exploration to collect the contexts about peers.

\textbf{Learning to Explore.} Exploration is a long-studied problem in single-agent and multi-agent reinforcement learning~\citep{hao2023exploration}. It is crucial to sufficiently explore the task space to find an optimal policy, but long-horizon and sparse reward properties may hinge effective exploration. To overcome the difficulties, several works~\citep{pathak2017curiosity, burda2018exploration, zhang2021noveld, badia2020never} introduce various intrinsic rewards measuring curiosity or dynamic error to boost exploration in single-agent reinforcement learning. Furthermore, there are several attempts at solving the multi-agent exploration problem. Maven~\citep{mahajan2019maven} proposes a shared latent space for hierarchical policy based on the value decomposition method for better exploration.  
CMAE~\citep{liu2021cooperative} introduces a shared common goal selected from restricted space to promote cooperative exploration.
Some works~\citep{iqbal2019coordinated, zheng2021episodic, jaques2019social} also try to extend the intrinsic reward to multi-agent exploration.
However, all of these works focus on how to effectively explore during the \textit{learning process} to obtain an optimal policy, whereas we focus on efficient exploration during \textbf{online interaction} with peers to identify their policies.
Our method leverages the auxiliary task to generate an intrinsic reward to boost the exploration strategy learning.

\vspace{-0.2cm}
\section{Method}

\subsection{Problem Formulation}

We formulate the underlying game of peer adaptation as a finite-horizon partially observable stochastic game (POSG)~\cite{hansen2004dynamic} $\left\langle\mathcal{I}, \mathcal{S},b0,\left\{\mathcal{A}_i\right\},\left\{\mathcal{O}_i\right\}, \mathcal{P},\left\{R_i\right\}\right\rangle$, where $\mathcal{I}$ is the finite set of all agents in the environment, $\mathcal{S}$ is the state space, $b0 \in \Delta_\mathcal{S}$ is the initial state distribution, $\mathcal{A}_i, \mathcal{O}_i, R_i$ is the action space, observation space, and reward function for agent $i$, and $\mathcal{P}$ is the transition function $P(s'|s, \mathbf{a})$.
Every episode of the game is guaranteed to terminate in finitely many time steps (finite-horizon).
Following prior work~\cite{fu2022greedy,gu2021online}, we consider the case with a single adaptive agent $\pi$ (ego agent)  and $m$ peer agents $\boldsymbol{\psi}$.
Our goal is to learn a policy $\pi_\theta$ that can adapt to various combinations of peer agents for the ego agent.

In peer adaptation, the ego agent will repeatedly interact with peer agents over $N_\text{eps}$ episodes of POSG to maximize the cumulative return.
Denoting the ego agent as agent $1$, we define the objective of \textbf{peer adaptation} over $N_\text{eps}$ episodes as the cumulative return over all episodes:
\begin{equation}
    \label{eq:meta_return_raw}
    \text{maximize} \quad \E [\sum_{n=1}^{N_\text{eps}} \sum_{t=1}^{T_n} r_{n, t}^1]
\end{equation}
where $r_{n, t}^1$ is the reward for the ego agent at time step $t$ of episode $n$, $T_n$ is the length of episode $n$. 
Our multi-episode objective poses significant challenges to learning adaptation policy, as a long horizon coupled with partial observability makes it dramatically harder to explore and exploit. Furthermore, it should be noted that different $N_\text{eps}$ induce different problem instances with corresponding optimal strategies.
A small $N_\text{eps}$ leaves little time for exploration, so the optimal policy may settle for an imperfect exploitative strategy, while a large $N_\text{eps}$ allows the ego agent to sacrifice some early return to explore and enable better exploitation later on.
See Appendix~\ref{sec:abl_horizon} for details.

\subsection{Context-aware Policy}

To optimize objective (Eq.~\ref{eq:meta_return_raw}), the ego agent needs to leverage its local trajectories of observations and actions, referred to as \textbf{context}, to infer the type and mind (e.g., belief, intention, desire) of its peer agents and take appropriate actions.
Formally, we denote the context for the ego agent as $C^1=\{\{o_{n,t}^1, a_{n, t}^1\}_{t=1}^{T_n}\}_{n=1}^N$, consisting of the ego agent's local observations and actions.
$T_n$ is the length of the $n$-th episode.
The context also includes the current episode $N$, which may be incomplete; in this case, $T_N$ is the current number of steps in episode $N$.

The number of episodes in the context $C^1$ may vary, and so do the lengths of the episodes in a single context. 
We build a \textbf{context encoder} $\chi$ parameterized by $\theta$ to encode contexts with varying sizes to a fixed-length vector $z^1 \in \mathbb{R}^m$:

\begin{equation}
    z^1:=\chi_\theta(C^1)= g_\theta\left(\frac{1}{N} \sum_{n=1}^N \frac{1}{T_n} \sum_{t=1}^{T_n} f_\theta(o^1_{n, t}, a^1_{n, t}) \right)
\end{equation}

where $f_\theta: \mathbb{R}^{|\mathcal{O}^1|} \times \mathbb{R}^{|\mathcal{A}^1|} \rightarrow \mathbb{R}^m, g_\theta: \mathbb{R}^m \rightarrow \mathbb{R}^m$ are MLPs, $\chi(\emptyset):= \mathbf{0}$.
With the context encoded by $\chi$, a \textbf{context-aware policy} $\pi_\theta(a|o, \chi_\theta(C))$ is trained to maximize the discounted reinforcement learning (RL) objective:
\begin{equation}
    \label{eq:meta_return}
    \text{maximize} \quad \E [\sum_{n=1}^{N_\text{eps}} \sum_{t=1}^{T_n} \gamma^{t'} r_{n, t}^1]
\end{equation}
where $\gamma \in (0, 1)$ is the discount factor, $t':=t+\sum_{n'=1}^{n-1} T_{n'}$ is the cumulative number of time steps until time step $t$ of episode $n$.
This objective is amenable to standard RL algorithms by concatenating $N_\text{eps}$ episodes together and computing the total returns, similar to the multi-episode return objective used in~\citep{xie2021learning}.

The encoder and the policy are jointly parameterized and optimized in an end-to-end manner using PPO~\citep{schulman2017proximal}.
By end-to-end training of both components, the encoder $\chi_\theta$ is always trained to extract the characteristics of peers from trajectories sampled by the \textit{current} policy $\pi_\theta$.
This eliminates the distribution mismatch potentially encountered by some prior works~\cite{fu2022greedy} that separate the learning of agent modeling and policy.

\subsection{Context-aware Exploration with Peer Identification}

For adaptation to unknown peers, the ego agent must first infer certain characteristics of the peer agents, e.g. strategies and preferences, to carry out the best response.
However, the inference process can be hindered in partially observable environments, as the behaviors of peer agents may not always be revealed in the ego agent's observation.
Peer adaptation thus requires a sophisticated exploration strategy, which is hard to learn directly from the original task reward.
For example, in the \textit{PO-Overcooked} environment, the ego agent can observe the peer agent's policy by going across the horizontal wall into the lower room.
However, this behavior can not yield any task reward immediately.
As a result, the learning policy will quickly converge during the initial stages of training to the local optimum strategy, i.e., not visiting the lower room.
This problem is further exacerbated by the requirement of adaptation across multiple episodes.
To properly adapt to the peers, the ego agent may need to perform an exploration behavior first to actively collect informative contexts for reasoning, identify the characteristics of the peers based on the observed context, and then carry out a specific response strategy.
Failure in any of these steps would lead to an overall failure and disrupt the learning of other steps.
As a result, it is extremely difficult to learn such strategies without any incentive for exploration.

To overcome this issue, we propose the maximization of the following mutual information objective:
\begin{equation}
\label{eq:mi}
        \begin{array}{cl}
    & I(\boldsymbol{\psi}, C^1) \\
    = & H(\boldsymbol{\psi}) - H(\boldsymbol{\psi} | C^1) \\
    = & H(\boldsymbol{\psi}) + E[\log P(\boldsymbol{\psi} | C^1)]
\end{array}
\end{equation}
where the randomness is taken over choice of peer agents $\boldsymbol{\psi}$, the policies of the ego agent $\pi_\theta$ and the peer agents $\boldsymbol{\psi}$, and the environment dynamics.
Intuitively, this objective encourages the ego agent to act in a way such that its context contains enough signals to identify the policies of the peer agents, leading to active exploration behaviors.
After thoroughly probing the peer agents, the ego agent can then adapt its policy to take the best response.
During training, we utilize a diverse peer pool $\boldsymbol{\Psi}$ from which peer agent policies $\boldsymbol{\psi}$ are sampled.
For this concrete peer distribution $\boldsymbol{\Psi}$, $H(\boldsymbol{\psi})$ is a constant with respect to parameters $\theta$.

Now, to compute $P(\boldsymbol{\psi} | C^1)$ and maximize it, we propose \textbf{peer identification} as an auxiliary task.
This task serves two purposes: the estimation of $P(\boldsymbol{\psi} | C^1)$ yields a better context representation, which is useful for downstream policy learning; the estimated $P(\boldsymbol{\psi} | C^1)$ can be added as an exploration reward to maximize Eq.~\ref{eq:mi} and promote exploratory behaviors.
We train an identifier $h_\theta: \mathbb{R}^m \rightarrow \Delta_{\boldsymbol{\psi}}$ to produce the posterior distribution of peer agents $h_\theta(\chi_\theta(C^1))$ given context $C^1$.
The identifier is jointly parameterized with the context encoder $\chi_\theta$.
The training of identifier $h_\theta$ depends on the parameterization of peer policies $\boldsymbol{\psi}$, which can be rule-based or parameterized by a neural network, etc.
For the general case without any knowledge or assumption about the peer pool $\boldsymbol{\Psi}$, we parameterize $\boldsymbol{\psi}$ as a tuple of indexes $(i_1, i_2, \ldots, i_m)$, where $\psi_{i_j} \in \boldsymbol{\Psi}=\{\psi_1, \psi_2, \ldots, \psi_{|\boldsymbol{\Psi}|}\}$ is the policy of the $j$-th peer agent.
In this case, the training loss for approximating the posterior distribution is given by
\begin{equation}
    L_\text{aux}(\theta) = \E_{\boldsymbol{\psi}, C^1} \left[\frac{1}{m} \sum_{j=1}^m \mathop{CE}(h_\theta(\chi_\theta(C^1))_j, i_j) \right]
\end{equation}
where $h_\theta(\chi_\theta(C^1))_j$ is the posterior categorical distribution for the index of peer agent $j$, $\mathop{CE}$ is the cross-entropy loss.
Optimizing $L_\text{aux}$ requires sampling paired peers and contexts $(\boldsymbol{\psi}, C^1)$.
We collect the contexts generated during RL training in a buffer to provide data for $L_\text{aux}$, so the identifier always stays on-policy.
See Section~\ref{sec:train_proc} for training details.

After estimating $P(\boldsymbol{\psi} | C^1)$, for maximizing the mutual information objective (Eq.~\ref{eq:mi}), we add an exploration reward based on the objective:
\begin{equation}
\label{eq:exp_reward}
    r^e := \frac{1}{m} \sum_{j=1}^m h_\theta(\chi_\theta(C^1))_{j, i_j}
\end{equation}
which is the estimated posterior probabilities of the actual peer agents, $r^e \in [0, 1]$.
We directly use the probability instead of the log version since the probability is bounded.

For encouraging the balance between exploring the peer agents and exploiting them to achieve task success, the final reward for the ego agent is computed as $r^1 = r + c \cdot r^e$, where $r$ is the original task reward and $c$ is a coefficient.
During training, we linearly decays $c$ from $c_\text{init}$ to $0$ in $M$ environment steps.
The exploration reward is not used during online adaptation, as the ground-truth peer identities are unknown.
See Appendix~\ref{sec:online_adp_app},\ref{sec:train_details} for details.

\subsection{Training Strategy} \label{sec:train_proc}

\begin{algorithm}[tb]
\caption{Training Procedure of PACE}\label{alg:train}
\begin{algorithmic}[1]
\REQUIRE Training peer pool $\boldsymbol{\Psi}$, context size $N_{\text{ctx}}$
\STATE Randomly initialize $\theta, t \gets 0$
\STATE $\forall \boldsymbol{\psi} \in \boldsymbol{\Psi}, C^1_{\boldsymbol{\psi}} \gets \emptyset$ \COMMENT{Initialize training contexts}
\STATE Initialize $|\boldsymbol{\Psi}|$ environments and reset to get $o^1_0$ for all $\boldsymbol{\psi}$
\WHILE{Maximum training step not reached}
\STATE $D \gets \emptyset$ \COMMENT{Initialize current training batch}
\WHILE{Current batch size not reached}
\FORALL{$\boldsymbol{\psi} \in \boldsymbol{\Psi}$} \STATE Step the env with $a^1_t \sim \pi_\theta(a|o^1_t, \chi_\theta(C^1_{\boldsymbol{\psi}}))$ and $\mathbf{a}^{-1}_t \sim \boldsymbol{\psi}$, obtain $r_t, o^1_{t+1}$
\STATE Compute $r^1_t$ using Eq.~\ref{eq:exp_reward} and $r^1_t = r_t + c \cdot r^e_t$
\STATE Put $(C^1_{\boldsymbol{\psi}}, \boldsymbol{\psi}, o^1_t, a^1_t, r^1_t)$ into $D$
\STATE Update $C^1_{\boldsymbol{\psi}}$ with $(o^1_t, a^1_t)$ \IF{$C^1_{\boldsymbol{\psi}}$ contains $N_\text{eps}$ complete episodes}
\STATE $C^1_{\boldsymbol{\psi}} \gets \emptyset$
\ENDIF
\ENDFOR
\STATE $t \gets t + 1$
\ENDWHILE
\STATE Update $\theta$ with PPO loss and $L_\text{aux}$ using $D$
\ENDWHILE
\end{algorithmic}
\end{algorithm}

\begin{figure*}[t]
    \centering
    \begin{subfigure}{0.33\textwidth}
        \includegraphics[width=\linewidth]{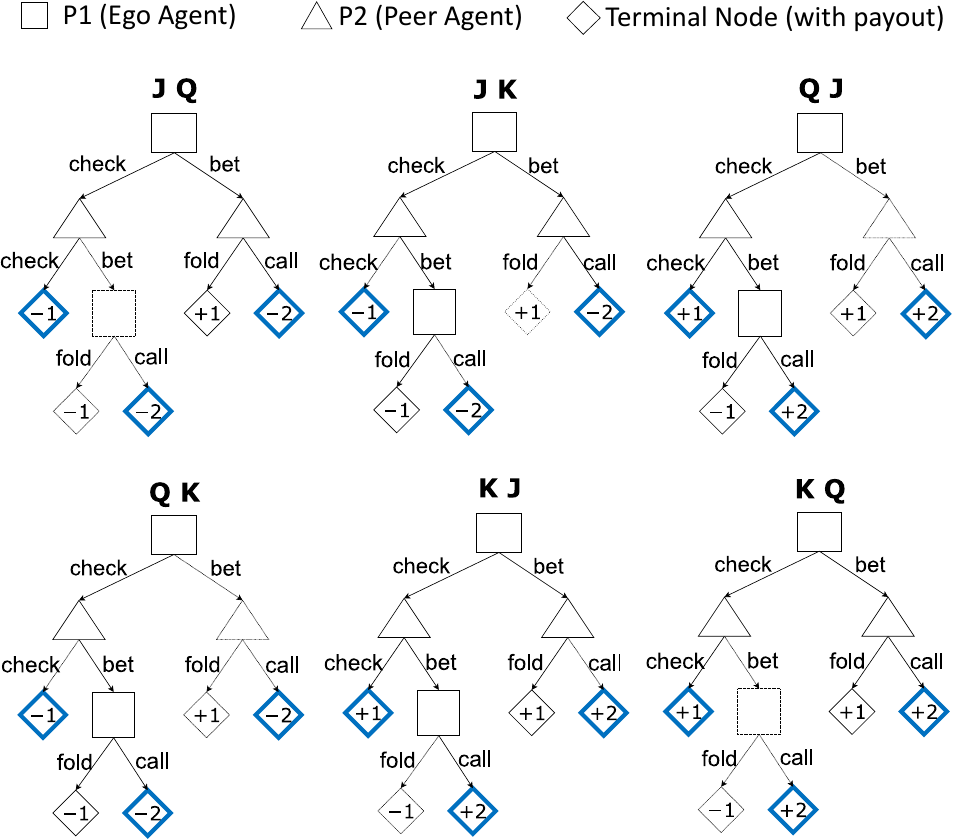}
        \caption{Kuhn Poker}
        \label{fig:kp_env}
    \end{subfigure}
    \begin{subfigure}{0.33\textwidth}
        \begin{subfigure}{0.49\linewidth}
        \includegraphics[width=\linewidth]{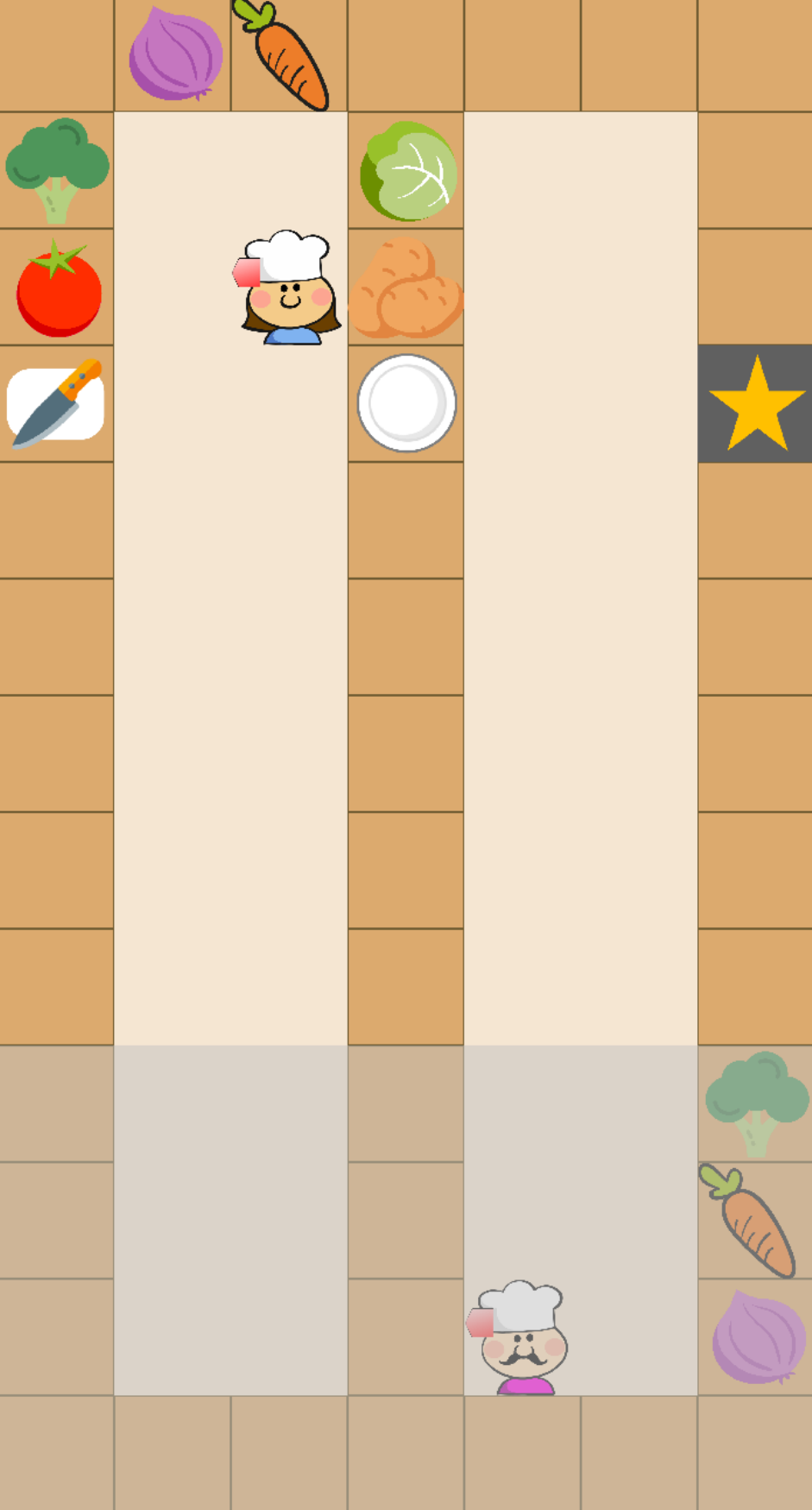}
                    \end{subfigure}
    \begin{subfigure}{0.49\linewidth}
        \includegraphics[width=\linewidth]{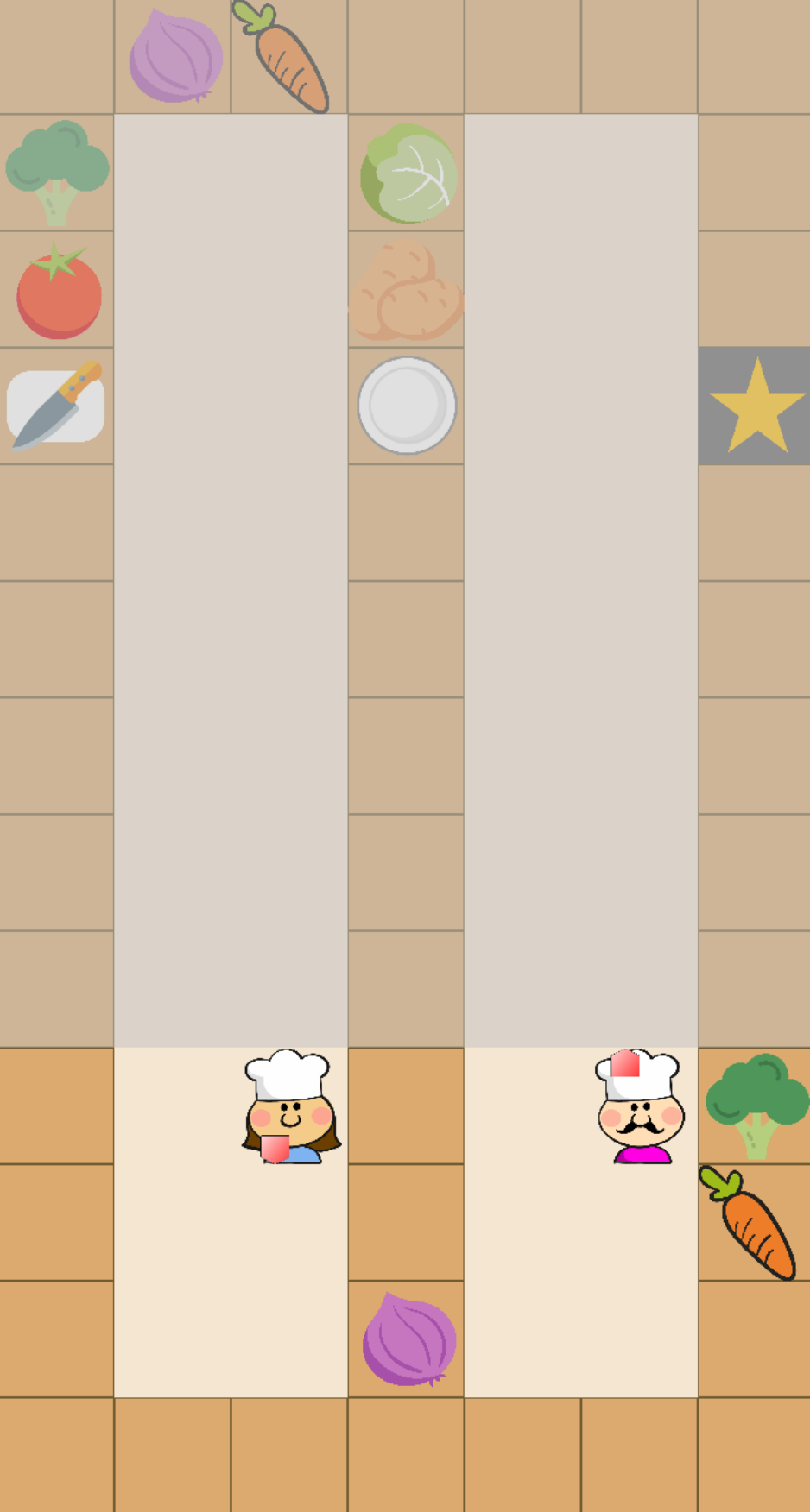}
                    \end{subfigure}
    \caption{PO-Overcooked}
    \label{fig:ovk_demo}
    \end{subfigure}
    \begin{subfigure}{0.3\textwidth}
        \includegraphics[width=\linewidth]{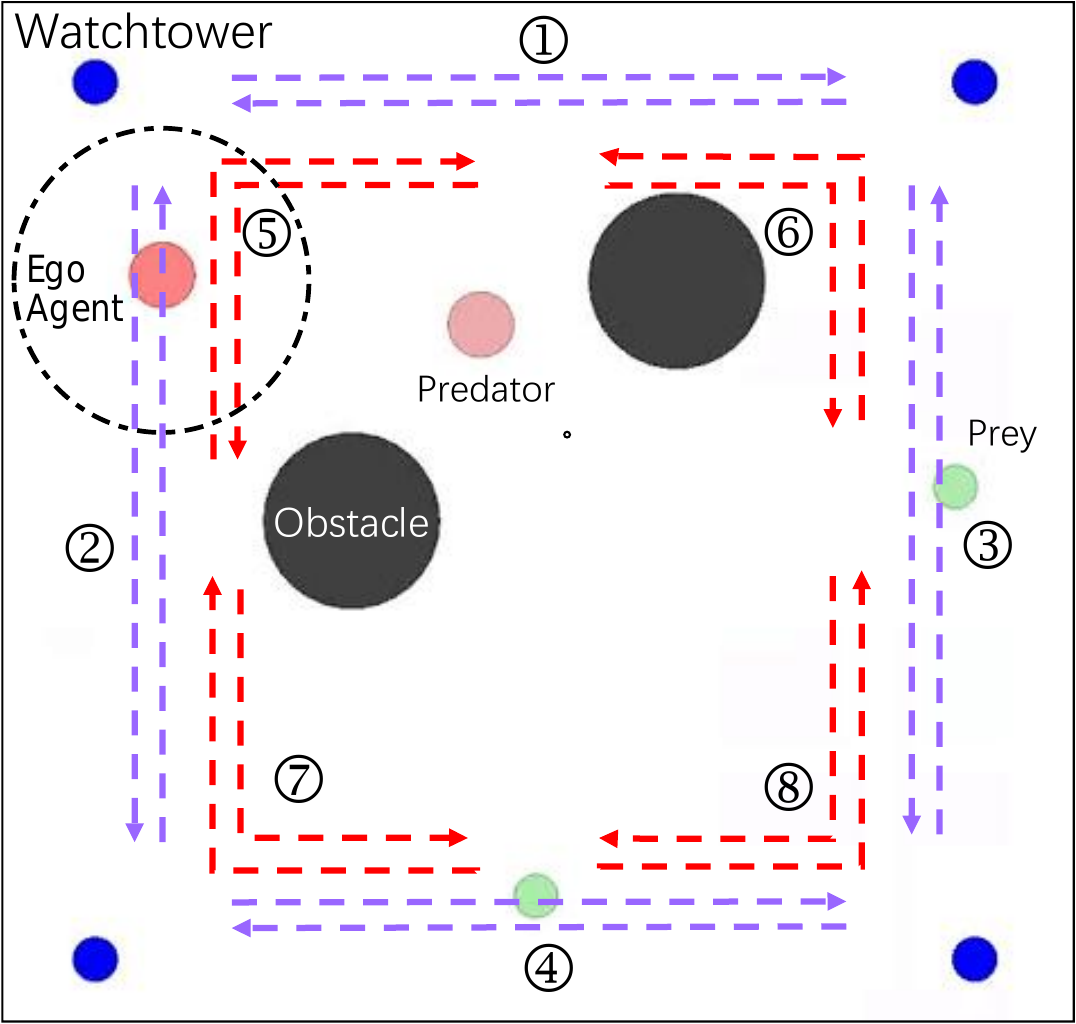}
        \caption{Predator-Prey-W}
        \label{fig:pp_env}
    \end{subfigure}
    \caption{Illustrations of Kuhn Poker (a), PO-Overcooked (b), and Predator-Prey-W (c). In (a), the hand of the peer agent is only revealed at showdowns (blue diamond nodes); in (b), the masked gray area indicates the unobserved area to the ego agent (the agent in the left room); 
        in (c), the ego predator can only have full observability during contact with the watchtowers (blue circles).}
\end{figure*}

We present our training algorithm that enables the ego agent to adapt to peers with markedly different strategies.
To accomplish this, we assume that a diverse peer distribution is available for training, from which we draw several policies as the training peer pool $\boldsymbol{\Psi}$.
The ego agent is trained to adapt to all the peers in the training peer pool $\boldsymbol{\Psi}$.

See Algorithm~\ref{alg:train} for pseudocode.
During training, we maintain an environment, a context, and a current observation for every tuple of peer agents $\boldsymbol{\psi}$ in the training peer pool $\boldsymbol{\Psi}$ (Line 2-3).
The context receives a new observation-action pair at every time step during policy rollout (Line 11) and gets cleared after reaching $N_\text{eps}$ episodes (Line 13).
During policy rollouts, the intrinsic exploration reward is generated and added to the task reward (Line 9).
After collecting a batch of data, we update the actor and critic for RL and the context encoder (Line 18).
The RL losses and $L_\text{aux}$ are computed with the same mini-batch and added together for optimization (Appendix~\ref{sec:train_details}).

\section{Experiments}
In this section, we conduct experiments in Kuhn Poker (Competitive), PO-Overcooked (Cooperative), and Predator-Prey-W (Mixed) to answer the following questions:
1) How well can PACE exploit the opponent in the competitive setting?
2) How well can PACE adapt to the partner in the cooperative setting?
3) How well can PACE perform in the presence of both kinds of peers?
4) Can PACE adapt to peers with sudden changes?
5) How do peer identification and the intrinsic reward influence learning and adaptation?
6) What does PACE learn in its latent space?

\subsection{Experiment Setup}

\begin{figure*}[t]
    \centering
    \begin{subfigure}{0.335\textwidth}
        \includegraphics[width=\linewidth]{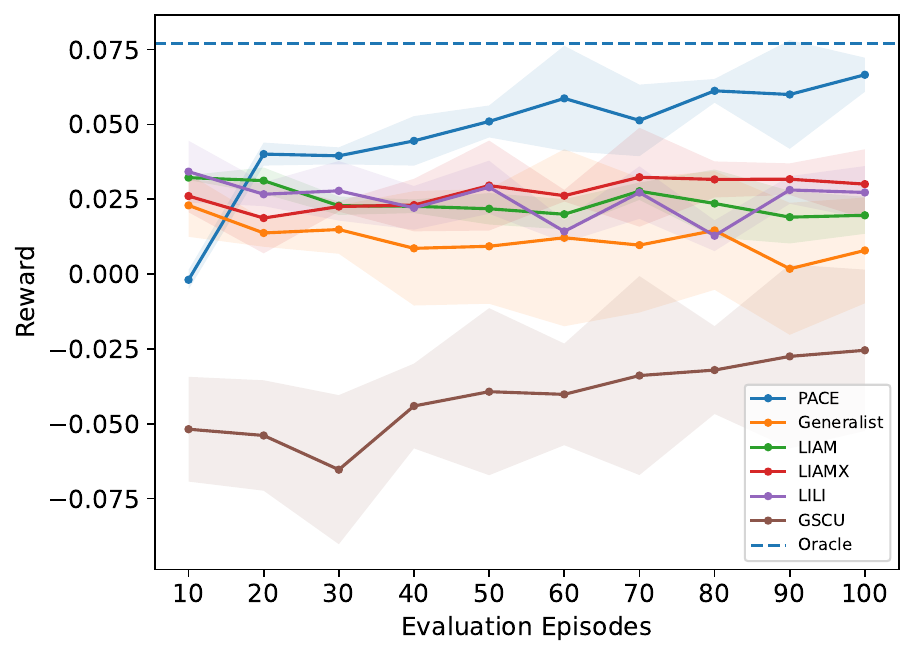}
        \subcaption{Kuhn Poker}
        \label{fig:main_kp}
    \end{subfigure}
                        \begin{subfigure}{0.315\textwidth}
        \includegraphics[width=\linewidth]{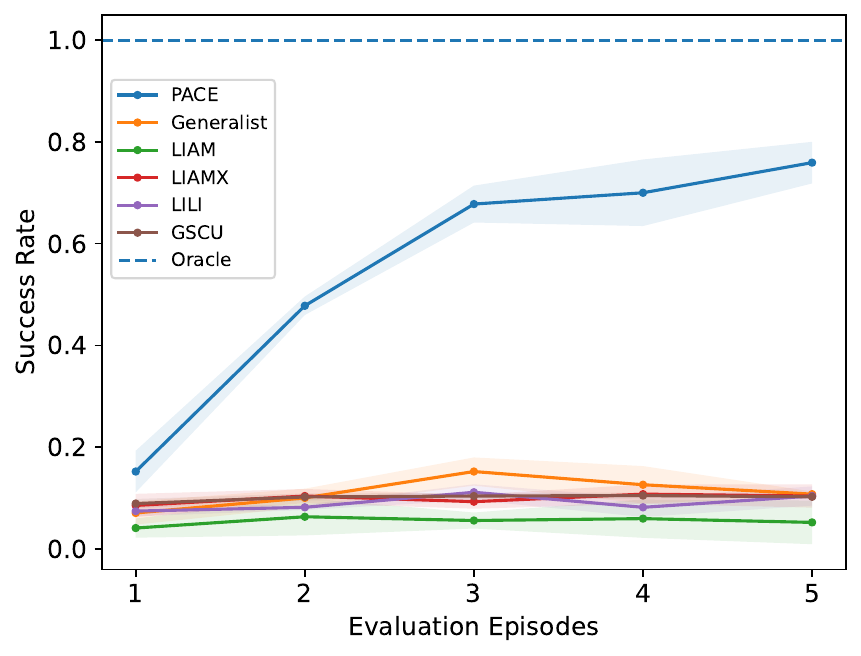}
        \subcaption{PO-Overcooked}
        \label{fig:main_ovk}
    \end{subfigure}
 	       \begin{subfigure}{0.33\textwidth}
        \includegraphics[width=\linewidth]{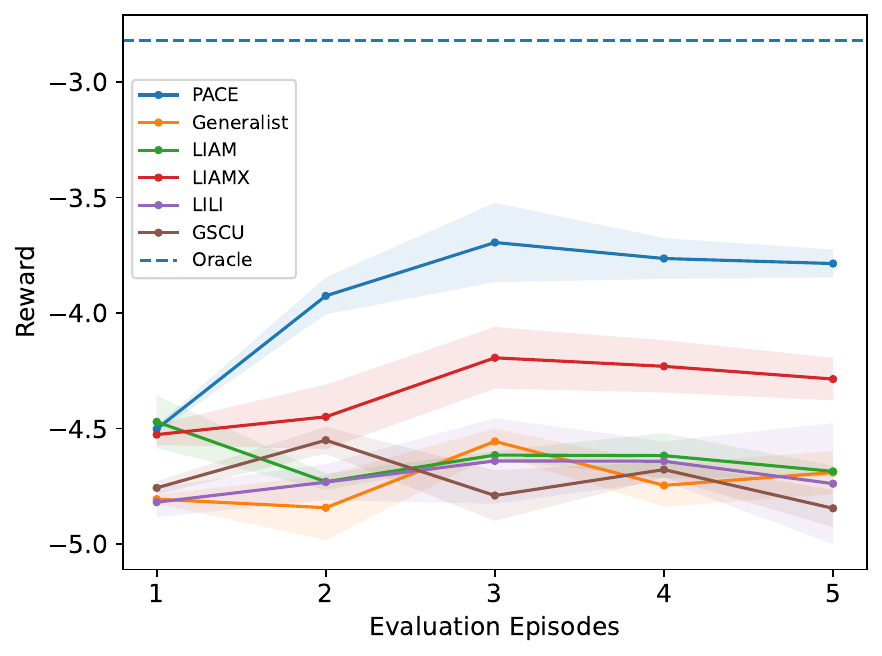}
        \subcaption{Predator-Prey}
        \label{fig:main_pp}
    \end{subfigure}
                                \caption{The online adaptation results on Kuhn Poker (a), PO-Overcooked (b), and Predator-Prey-W(c). PACE continuously improves over the whole online adaptation process, outperforming baselines in all environments. In particular, PACE is the only agent capable of adaptation in the PO-Overcooked environment. Oracle denotes the best responses designed separately for every peer in the test pool.}
   \end{figure*}

To demonstrate the validity of PACE, we conduct online adaptation experiments in three commonly used environments in the field of MARL: Kuhn Poker~\citep{kuhn1950simplified}, PO-Overcooked~\citep{carroll2019utility}, and Predator-Prey-W~\cite{lowe2017multi}.
The environments encompass a wide spectrum of scenarios, characterized by diverse aspects such as cooperative, competitive, and mixed settings; partial observability; multiple peer agents; as well as short and long time horizons.
We use the average episodic rewards or success rates over $N_\text{eps}$ episodes as the evaluation metric.

\textbf{Kuhn Poker}~\citep{kuhn1950simplified}. This is a simplified two-player poker game involving a deck of three cards and at most three rounds.
The game trees for every possible assignment of hand cards are shown in Figure~\ref{fig:kp_env}.
Both players place an ante of 1 before any action, and each player is dealt one private card initially.
The left card is for P1 and the right card is for P2 in Figure~\ref{fig:kp_env}. 
The players, P1 and P2, take turns to decide whether to Bet (Call) or Check (Fold).
The game ends when one of the players unilaterally folds and forfeits the pot to the other player.
If neither of the players folds, the game ends in a showdown (Figure~\ref{fig:kp_env}, blue diamond nodes), where the hands of the players are revealed to each other to decide the winner.
In this paper, the ego agent is P1 while the peer agent plays P2.
The peer agent pool is generated similarly as previous work~\citep{southey2009effective}, which parameterizes the P2 policy space with two parameters after eliminating dominated strategies.
To determine the peer policy, the ego agent needs to observe the hand card of the peer.
However, as the hand card is only publicly revealed during the showdown, the ego agent should decide whether to pursue a showdown and reveal information even when it may bring undesirable returns for the current episode.

\textbf{Partially Observable Overcooked (PO-Overcooked)}. Overcooked~\citep{carroll2019utility} is a collaborative cooking game where agents, acting as chefs, work together to complete a series of sub-tasks and serve dishes.
To add to the challenge and promote diverse policy behaviors, we provide a partially observable multi-recipe version of Overcooked based on~\citep{charakorn2023generating}, shown in Fig~\ref{fig:ovk_demo}.
Our scenario features a total of 6 kinds of ingredients and 9 recipes.
The game environment includes a series of counters that divide the room, compelling the agents to collaborate by passing objects, including ingredients and plates, over the counter.
The ego agent is the left agent in charge of making dishes while the peer agent delivers them at the right.
There is also a horizontal wall across the room, blocking the line of sight of the agents, forcing them to move across the wall to see the other side.
In Figure~\ref{fig:ovk_demo}, the left figure shows the ego agent in the upper room and can't see the objects in the lower room, unless it goes across the horizontal wall as in the right figure.

To generate a diverse peer pool, we adopt a rule-based approach by constructing a collection of peer policies as the right agent, each representing a specific preference for ingredients and recipes.
The rule-based agents are designed to take and serve only the ingredients and dishes that match their preferences.
Existing approaches~\citep{strouse2021collaborating, charakorn2023generating, lupu2021trajectory} primarily employ Reinforcement Learning algorithms in conjunction with diversity objectives to train policies that exhibit a wide range of behaviors.
In this work, we leverage these preference-based policies to capture more human-like behavior within the game (Appendix \ref{sec:peer_pool_ovk_app}).
The ego agent should explore the peer's preferences and determine its preferred recipe before making the target dish.

\textbf{Predator-Prey with Watchtowers (Predator-Prey-W)}. We employ the predator-prey scenario in Multi-agent Particle Environment~\cite{lowe2017multi} with both collaborative and competitive elements.
To increase the difficulty of exploration, we introduce partial observability and \textit{watchtowers} to the environment.
While the observation of the ego agent is restricted to a small radius around the agent (Figure~\ref{fig:pp_env}, black dotted circle), it may choose to visit the watchtowers around the map, which restore full observability for it during contact but yield no immediate rewards.
To construct diverse behaviors, prey policies in the peer pool are rule-based policies with different preferences on the sequences to reach landmarks, while every predator policy prefers to chase specific prey.
An adaptive policy for the ego agent involves discovering which trajectory each prey takes and which prey each predator is chasing, and following the correct prey.

\textbf{Baselines.} We choose the following baselines to validate the adaptation and peer modeling capability of PACE.
1) \textit{GSCU}~\citep{fu2022greedy} trains a conditional policy conditioned on a pre-trained opponent model. Notably, GSCU assumes the availability of peer observations and actions after the end of an episode, which is not necessary for PACE.
2) \textit{LIAM}~\citep{papoudakis2021agent} models the opponent's observation and action as an auxiliary task under partially observable settings.
3) \textit{LIAMX} is a variant of LIAM with cross-episode contexts.
4) \textit{LILI}~\citep{xie2021learning} models the transitions observed by the ego agent using the last episode, implicitly encoding the opponent as environment dynamics for the ego agent.
5) \textit{Generalist} is a plain recurrent policy with access to cross-episode contexts.

\subsection{How well can PACE exploit the opponent in the competitive setting?}

In Kuhn Poker, we randomly sample 40 \textit{P2} policies from the parameterized policy space as the training pool.
We also sample another 10 different policies for online adaptation testing.
This testing procedure spans $N_\text{eps}=100$ episodes.
The average rewards every 10 episodes during the online adaptation are presented in Figure \ref{fig:main_kp}. Standard deviations are reported over 3 training seeds. The average rewards over all 100 episodes are reported in the Appendix (Table \ref{tab:main}).

As is shown in Figure~\ref{fig:main_kp}, PACE continuously improves its rewards and outperforms the baselines.
This demonstrates the effectiveness and efficiency of the learned policy in adapting to the unknown opponents.
In particular, we note that the PACE agent also explicitly opts to explore the peer's strategy by using a more aggressive strategy for the first $10$ episodes.
During this time, the game enters showdown at a higher rate of $\sim 0.64$ for the PACE agent, so it gets to see the peer's hand more frequently and obtains more information about the peer's strategy.
While this leads to certain short-term losses in rewards, the overall performance is greatly improved to $0.047$, while the best baseline fetches $0.027$ (Table~\ref{tab:main}).
In comparison, the showdown rate for LILI holds steady at $\sim 0.60$ over all of the $100$ episodes of interaction.
GSCU also improves along the online interactions but fails to reach a satisfying level of rewards within the testing time horizon, due to a low starting point.
During online adaptation, GSCU may at times use a ``conservative policy" that is the Nash Equilibrium (NE) policy in Kuhn Poker.
This NE policy does not actively try to exploit its opponent, leading to a relatively unsatisfactory performance at first.
Other baselines mainly fluctuate around the initial performance, showing that it is hard to make use of the context and explore peer strategies without proper guidance.

\vspace{-0.2cm}
\subsection{How well can PACE adapt to the partner in the cooperative setting?}
\vspace{-0.1cm}

The PO-Overcooked environment poses a significant challenge for adaptation in coordination.
To cook a meal, the ego agent in the left room has to work together with the peer agent in the right room.
Note that the agent can not go to the other room, forcing the collaboration.
The context consists of a small number of episodes ($N_\text{eps}=5$), each lasting for dozens of steps.
However, as the peer agent only touches ingredients and dishes within its preference and ignores everything else, most of the context contains little information about its true preference.
This requires the ego agent to actively perform exploratory actions.
We sample 18 policies from the training pool and another 9 policies from the testing pool.

Figure \ref{fig:main_ovk} shows that PACE is the only agent that can adapt to the peer in the PO-Overcooked environment, achieving an average success rate of $0.553$, while all of the baselines fail with success rates of around $0.1$. Standard deviations are reported over 3 training seeds. The average rewards are reported in Table \ref{tab:main} in the Appendix.
Specifically, while GSCU can adapt to its peers and keep improving in Kuhn Poker, it fails in PO-Overcooked due to the lack of an effective exploration strategy.
The conservative policy of GSCU in PO-Overcooked is a generalist policy that rarely serves the preferred dish.
Consequently, the peer stands still for most of the time, revealing little information for GSCU to model.
The results also demonstrate that the context encoder $\chi_\theta$ can efficiently summarize long-term contexts and capture only the useful portion.

\vspace{-0.2cm}
\subsection{How well can PACE perform in the presence of both kinds of peers?}
\vspace{-0.1cm}
We further validate the performance of PACE in Predator-Prey-W with both cooperative and competitive peers.
In Predator-Prey-W, the ego agent is a predator seeking to collaborate with a peer predator and catch two peer prey.
To achieve this, the ego agent must first determine the trajectories of the prey and which prey the peer predator prefers, then track the other prey.
Each of these steps is hard in a partially observable environment, which restricts the observation of each agent to a small radius around the agent.
To mitigate this restriction, the ego agent can visit watchtowers around the corners of the map to gain full observability during contact with the watchtowers.
We sample $16$ combinations of training prey and predator policies as the training pool and $24$ separate combinations with unseen prey policies for online adaptation.

We report the online adaptation performance over $N_\text{eps}=5$ episodes in Figure \ref{fig:main_pp} and average performance in Table \ref{tab:main}.
Standard deviations are reported over 3 training seeds.
PACE achieves higher rewards than all the baselines over the whole adaptation procedure.
The only baseline capable of some adaptation is LIAMX, demonstrating that cross-episode contexts with auxiliary objectives can indeed improve performance.
However, LIAMX still underperforms PACE, as the objective of LIAMX is simply to supervise the encoder without altering policies, while PACE generates additional rewards to maximize the mutual information objective in Eq.~\ref{eq:mi}.
GSCU similarly fails to adapt as in PO-Overcooked.
The smaller number of episodes in both environments may also contribute to the underperformance of GSCU, as GSCU runs variational inference and adjusts its greedy policy only once after each episode.

\vspace{-0.2cm}
\subsection{Can PACE adapt to peers with sudden changes?} \label{sec:dyn_peer}
\vspace{-0.1cm}
\begin{table}[tb]
  \caption{The average success rates of PACE and LIAMX when adapting to suddenly changed peers in PO-Overcooked. PACE (*) is the success rate with stationary peers for reference.}
    \vspace{-0.2cm}
  \label{tab:dyn_ovk}
  \centering
  \begin{center}
  \scalebox{0.83}{
    \begin{tabular}{ccccc}
        \toprule
        PACE (*) & PACE & Generalist & LIAMX \\
        \midrule
        \underline{0.553 $\pm$ 0.029} & \textbf{0.435 $\pm$ 0.027} & 0.120 $\pm$ 0.007 & 0.116 $\pm$ 0.004 \\
        \bottomrule
    \end{tabular}
  }
  \vspace{-0.3cm}
  \end{center}
\end{table}

In addition to adapting to stationary peers, in this section, we demonstrate that PACE can detect and adapt to changes in peer policies by conducting a sudden-change experiment in PO-Overcooked.
Over $10$ adaptation episodes, we use a single tuple of peer agents in the first $5$ episodes, then switch to another tuple of peer agents for the last $5$ episodes.
However, the timing of the peer switch is unknown to the ego agent.
To detect the change in peer strategies, we use a heuristics strategy based on the reward observed by the ego agent. See Appendix \ref{sec:dyn_peer_app} for details.
The ego agent clears its context and restarts exploration after detecting such changes.
As shown in Table \ref{tab:dyn_ovk}, the PACE agent successfully detects the peer change and adapts accordingly with small performance degradation compared to the static case.
The baselines learn policies that are unresponsive to peer policies.
As a result, the baselines also fail to adapt in the sudden-change setting, achieving consistently low success rates.

\subsection{How do the auxiliary task and reward influence learning and adaptation?}
\begin{table}[tb]
  \caption{The average success rates of PACE and ablations in \textit{PO-Overcooked}.}
    \vspace{-0.1cm}
  \label{tab:abl_ovk}
  \centering
  \begin{center}
      \begin{tabular}{ccc}
        \toprule
        PACE & PACE-reward & PACE-reward-aux \\
        \midrule
        \textbf{0.553 $\pm$ 0.029} & 0.173 $\pm$ 0.016 & 0.143 $\pm$ 0.020 \\
        \bottomrule
    \end{tabular}
      \vspace{-0.3cm}
  \end{center}
\end{table}
Here we perform ablation experiments to examine the effect of the auxiliary task and reward on the training process and final convergence.
Table~\ref{tab:abl_ovk} contains the average success rates over the online adaptation procedure for PACE and ablations in PO-Overcooked.
PACE-reward-aux is PACE with neither the auxiliary task nor the exploration reward, while PACE-reward ablates the reward and retains the task.
It can be observed that the performance drops severely from $0.553$ to $0.173$ after removing the auxiliary reward, indicating that the reward is critical for performance.
Without the exploration reward, the policy quickly converges to the local optimum of not visiting the lower room.
Further removing the auxiliary task also hurts performance.
In comparison, the full PACE agent visits the lower room about once on average in an online adaptation procedure, as shown in Figure~\ref{fig:abl_ovk_train} in the Appendix.

\subsection{What does PACE learn in its latent space?}

\begin{figure}[t]
  \begin{subfigure}{0.24\textwidth}
          \includegraphics[width=\linewidth]{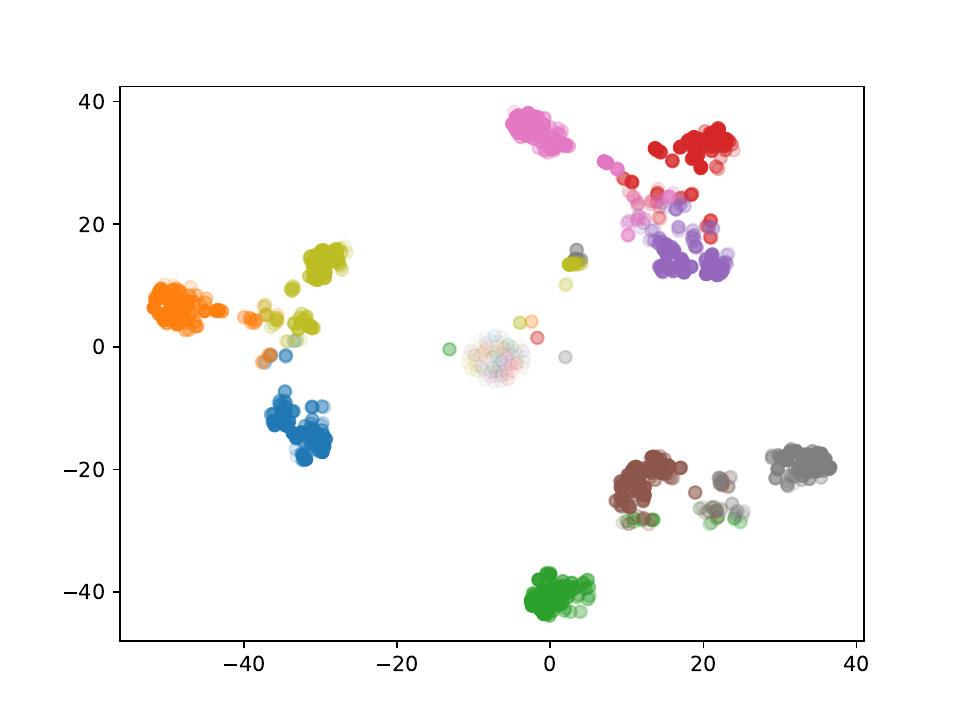}
     \caption{PACE}
     \label{fig:pace_latents}
 \end{subfigure}
 \begin{subfigure}{0.235\textwidth}
     \includegraphics[width=\linewidth]{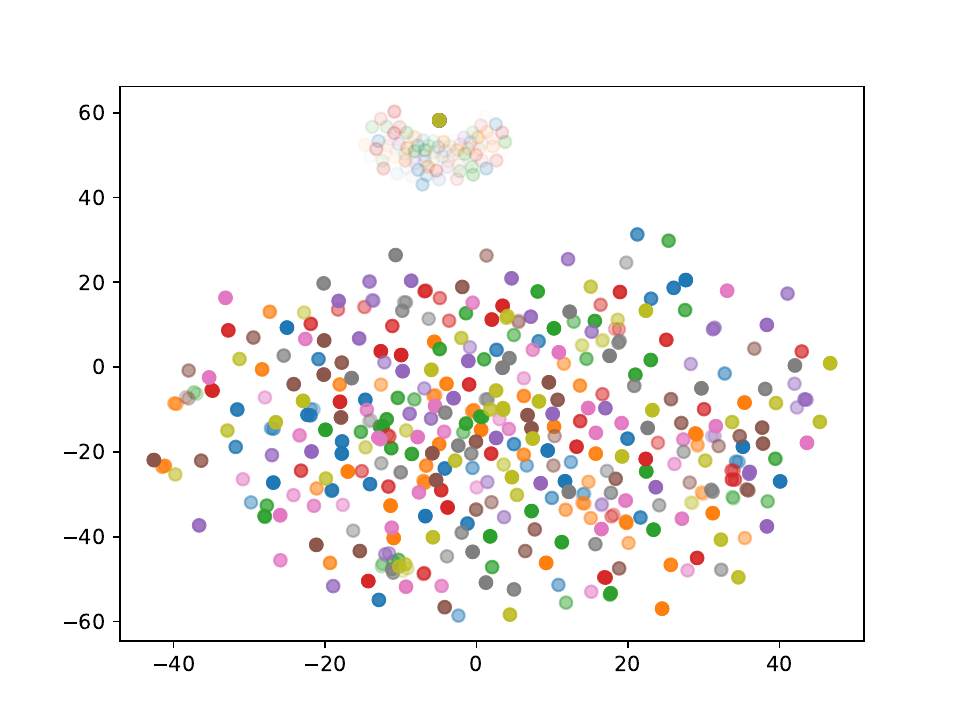}
     \caption{LILI}
     \label{fig:lili_latents}
 \end{subfigure}
 \vspace{-0.3cm}
 \caption{The t-SNE plot of the latent embeddings produced by the PACE (a) and LILI (b) encoder in \textit{PO-Overcooked}. Each color indicates a specific testing peer, while the shades of color denote the time order during adaptation.}
  \vspace{-0.3cm}
 \label{fig:latents}
\end{figure}

We visualize the latent space of PACE and compare it with baselines by collecting the generated context embeddings during online adaptation and projecting them into 2 dimensions using t-SNE.
Figure \ref{fig:latents} is a visualization of latent embeddings generated by the encoder of PACE and LILI~\citep{xie2021learning} in PO-Overcooked.
Each color corresponds to one of the 9 testing peers, while the shades of color denote the chronological order during adaptation.
It can be seen that at the beginning of adaptation, both PACE and LILI do not know the preferences of the peer agents, as the context is empty.
This is indicated by the light cluster in the middle of Figure~\ref{fig:pace_latents} and at the top of Figure~\ref{fig:lili_latents}.
However, with the accumulation of more interaction contexts, PACE successfully probes the peers and distinguishes between the contexts of different peers, as the latent first split into three broad clusters, then into nine small clusters corresponding to each test peer.
In contrast, the LILI latent scatter around the space with no discernible pattern, indicating that the encoder is unaware of the peer's identity.

\vspace{-0.3cm}
\section{Conclusion}
\vspace{-0.2cm}
In this paper, we propose fast peer adaptation with context-aware exploration (PACE), a method for training agents that explore and adapt to unknown peer agents efficiently.
Autonomous agents in the real world observe and adapt to the behaviors of their peers, whether to facilitate cooperation or exploitation, namely \textbf{peer adaptation}.
To achieve this goal, agents need to strike the right balance between exploration and exploitation, recognizing the peer policies before taking the appropriate response.
PACE leverages peer identification as an auxiliary task to guide context encoder learning and generate exploration rewards for the agent.
With a diverse pool of training peers, PACE trains a context-aware policy to maximize both the original task reward and the exploration reward.
The policy explores the peers when it is uncertain about the best response, and exploits the best response otherwise.
We conduct experiments in Kuhn Poker, PO-Overcooked, and Predator-Prey-W, three popular MARL environments covering a wide range of properties.
Experimental results confirm that the PACE agent can efficiently explore the peers and adapt its policy based on the context, achieving good performance in competitive, collaborative, and mixed scenarios.

\textbf{Limitation and Future Work.}
There are certain limitations to PACE.
The algorithm requires a diverse peer pool for training, which is essential for training an effective adaptive agent.
Generating more complex and sophisticated strategies for the training peer pool could further enhance the performance of PACE.
Currently, our experiments involve the same number of agents in testing as in training.
Addressing scenarios where the number of agents may change between episodes is another important future direction.
Alternative auxiliary tasks may also benefit policy learning.
Moreover, extending our approach to heterogeneous multi-agent games~\citep{wang2024romat, pan2022mate} and embodied multi-agent scenarios~\citep{wu2022grasparl, zhong2023rspt, chen2023bi, zhong2024empowering}. The research on the peer adaptation of embodied agents in complex 3D environments~\citep{qiu2017unrealcv} could open up new possibilities and challenges for multi-agent learning.
Furthermore, although a key goal of multi-agent learning is to build agents that can interact with humans, we use no human peers in this work.
An important direction is to conduct human studies to evaluate how PACE agents can interact with human peers in various settings.
This would require addressing issues such as human factors, ethical considerations, and user feedback.

\section*{Acknowledgements}
This work was supported by the National Science and Technology Major Project (MOST-2022ZD0114900), China National Post-doctoral Program for Innovative Talents (No. BX2021008), and Qualcomm University Research Grant.

\section*{Impact Statement}

Peer adaptation is a fundamental problem in the area of multi-agent reinforcement learning, as collaboration and competition are everywhere in the real world.
Research in this area facilitates these interactions for autonomous agents in the future and improves their efficiency.
PACE can also help with the interactions between humans and autonomous agents, like home care robots adapting to the needs of the elderly.
However, like other machine learning techniques, PACE faces the risk of misuse, especially in competitive settings.
For example, malicious actors may use it to engage in illegal activities and circumvent law enforcement.
Active mitigation efforts are required before PACE can be deployed in the real world.

\bibliography{main}
\bibliographystyle{icml2024}

\newpage
\appendix
\onecolumn

\begin{table}[tb]
  \caption{The average episodic performance of PACE and baselines. The Oracle best response performance is shown for reference.}
  \label{tab:main}
  \centering
  \begin{center}
  \scalebox{0.86}{
  \begin{tabular}{cccc}
    \toprule
        \multirow{2}{*}{Methods} & Kuhn Poker & PO-Overcooked & Predator-Prey-W \\
    & (Reward) & (Success rate) & (Reward) \\
    \midrule
    Oracle & \textcolor{gray}{0.077} & \textcolor{gray}{1.0} & \textcolor{gray}{-2.82} \\
    PACE & \textbf{0.047 $\pm$ 0.004} & \textbf{0.553 $\pm$ 0.029} & \textbf{-3.93 $\pm$ 0.04} \\
    Generalist & 0.012 $\pm$ 0.016 & 0.111 $\pm$ 0.016 & -4.73 $\pm$ 0.03 \\
    LIAM & 0.024 $\pm$ 0.001 & 0.054 $\pm$ 0.027 & -4.62 $\pm$ 0.02  \\
    LIAMX & 0.027 $\pm$ 0.008 & 0.099 $\pm$ 0.009 & -4.34 $\pm$ 0.08 \\
    LILI & 0.025 $\pm$ 0.001 & 0.090 $\pm$ 0.007 & -4.71 $\pm$ 0.09 \\
    GSCU & -0.041 $\pm$ 0.021 & 0.100 $\pm$ 0.005 & -4.72 $\pm$ 0.05 \\
    \bottomrule
  \end{tabular}
  }
  \end{center}
\end{table}

\begin{figure*}[t]
    \centering
    \begin{subfigure}{0.45\textwidth}
        \includegraphics[width=\linewidth]{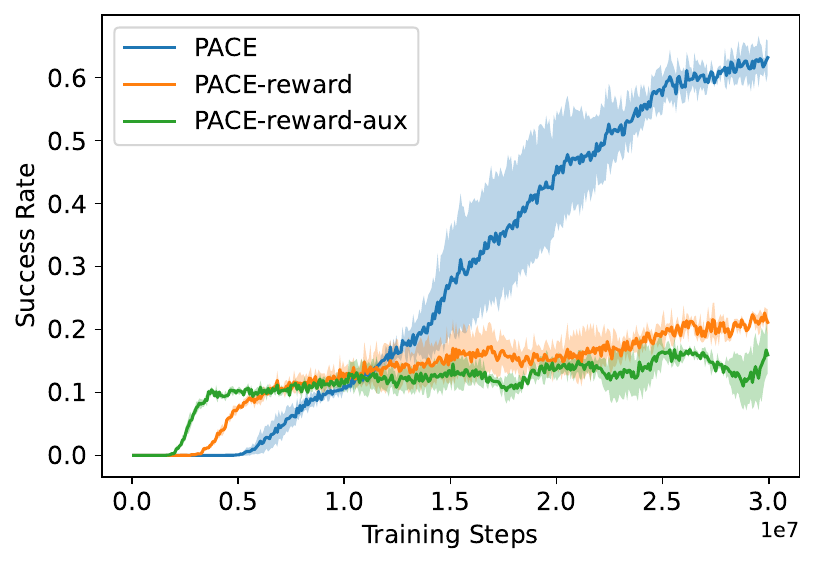}
    \end{subfigure}
    \begin{subfigure}{0.45\textwidth}
        \includegraphics[width=\linewidth]{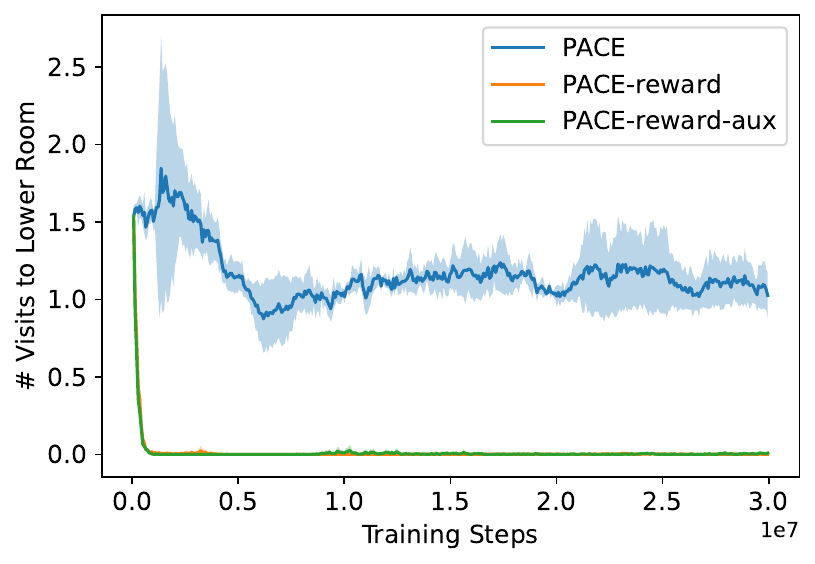}
    \end{subfigure}
    \caption{Training curves of PACE and ablations on \textit{PO-Overcooked}. PACE has a higher success rate (a) and explores the lower room about once per adaptation procedure (b), while ablations fail to explore and can't cooperate properly.}
    \label{fig:abl_ovk_train}
\end{figure*}

\section{Environment Details}

\subsection{Kuhn Poker}
Kuhn Poker is a simplified two-player (P1 \& P2) poker game~\citep{kuhn1950simplified}. The game involves a deck of three playing cards, where each player is dealt one card. The cards are ranked (from lowest to highest) Jack, Queen, and King. There are no suits in Kuhn poker, only ranks. The action is restricted to bet and pass, different from No Limit Texas Hold'em, which supports multi-round raising. The game proceeds as follows:
\begin{enumerate}
    \item Both players put one ante (chip) into the pot.
    \item Each player is dealt with one card from the deck. The remaining card is unseen by both players.
    \item After the deal, P1 is the first to take action, choosing to bet 1 chip or pass.
        \begin{itemize}
            \item If P1 chooses to bet, then P2 can bet (call P1's bet and the game ends in a showdown) or pass (fold and forfeit the pot).
            \item If P1 chooses to pass, then P2 can pass (check and the game ends in a showdown) or bet.
            \item If P2 bets after P1 passes, P1 should choose to bet (call P2's bet, and the game ends in a showdown) or pass (fold and forfeit the pot).
        \end{itemize}
\end{enumerate}

In this paper, we focus on learning the adaptation strategy of P1 against P2, and the peer plays as P2. Following the strategy simplification approach introduced by Southey et al.~\citep{southey2009effective}, we eliminate obviously dominated policies for P2. For example, P2 never bets with Queen after P1 checks, because P1 will always fold with Jack and always call with King. The whole simplified game tree can be found in the paper~\citep{southey2009effective}. This simplification allows us to parameterize the P2 policy using two parameters ($\xi,\eta$) within the range of 0 to 1. $\eta$ is the probability of betting Queen after P1 bets. $\xi$ is the probability of betting Jack after P1 passes. Consequently, the entire policy space of P2 can be divided into six sections, each corresponding to the best response from P1.

\textbf{Observation Space.} The agents in the game observe a state represented by a 13-dimensional vector, consisting of 3 one-hot vectors. 
The first one-hot vector is a 7-dimensional representation of the current stage in the game tree.
The second and third one-hot vectors, each of 3 dimensions, represent the hand card of the ego player and the opponent. If it has not come to a showdown stage, the opponent hand is always represented by an all-zero vector.

\textbf{Action Space.} As stated above, each player can only choose to bet or pass, so the action space is a discrete space with 2 actions.

\textbf{Reward.} The reward is not directly determined by the pot itself. Instead, it is calculated based on the chips present in the pot minus the chips contributed by the winner. For the loser, the reward is the negative value of the winner's reward. If the game ends in a showdown, the player holding the highest-rank card wins the pot. If no player bets, then the pot is 2, so the reward is $\pm$1. Otherwise, the pot is 4 (one player bets, and the other one bets thereafter), so the reward is $\pm$2. If the game ends due to one player forfeiting the pot, then the other player wins the pot of 3 chips, so the reward is $\pm$1. 

\subsection{PO-Overcooked}

PO-Overcooked is a collaborative cooking game where players take on the roles of chefs working together to complete various sub-tasks and serve dishes~\citep{carroll2019utility}.
In this paper, we introduce a more complex Multi-Recipe version, which builds upon the modifications by Charakorn et al.~\cite{charakorn2023generating}. 
Specifically, we add two extra ingredients, potato, and broccoli, and correspondingly more recipes to increase the challenge and encourage diverse policy behaviors. 
The game scenario involves a total of 6 ingredients (Tomato, Onion, Carrot, Lettuce, Potato, and Broccoli) and 9 recipes. 
Notably, the game environment features a counter that divides the room, necessitating collaboration between chefs as they pass objects such as ingredients and plates back and forth over the counter. 
To serve a dish, the necessary ingredients should be first taken to the cut board and chopped. 
After all the required ingredients are chopped and put onto a plate, the dish needs to be carried to the delivery square to finish the task.
Furthermore, we add partial observability to the game, separating the game scene horizontally into an upper room and a lower room.
Each agent can only see objects in the same room as itself.

\textbf{Observation Space.} The observation is a 105-dimensional vector. It consists of multiple features, including position, direction, holding objects, front objects, and so on.
There is a flag for each relevant object that indicates its visibility to account for partial observation.

\textbf{Action Space.} Each agent can choose from a discrete space with 6 actions: move left/right/up/down, interact (with objects), and no-op (take no action).

\textbf{Reward.} PO-Overcooked is a fully cooperative game, so all agents share the reward. There are three types of rewards in the game. The first is the interactive reward. Each agent receives a reward of 0.5 if an object is interacted with by an agent. Note that repeated interactions with the same object do not accumulate additional rewards. The second is progress reward. Each agent receives a reward of 1.0 when the state of a recipe progresses. For example, if a chopped carrot is placed into a plate, transitioning the recipe state from "chopped carrot" to "carrot plate," each agent is rewarded. The third is complete reward. When a dish satisfying a recipe is served to the deliver square, each agent receives a reward of 10.0.

\subsection{Predator-Prey-W}

Here we introduce an environment with multiple peers, where some peers cooperate with the ego agent and others compete with it.
We use a modified version of the predator-prey scenario from the Multi-agent Particle Environment (MPE)~\citep{lowe2017multi} commonly used in the MARL literature.
As illustrated in Figure~\ref{fig:pp_env}, the environment features two predators (red circles, the darker one of which is the ego agent), two prey (green circles), and multiple landmarks (grey and blue circles).
The predators are tasked with chasing the prey while the prey escapes from the predators.
Furthermore, the predators are required to collaborate such that all of the prey are covered by predators (see the Reward section below).
Each episode lasts for at most $40$ steps.
If all the prey have been touched by predators, the episode terminates immediately.

To make the task harder, we additionally introduce partial observability and four watch towers (blue circles, corners of the figure).
The ego agent can only observe agents and landmarks within its observation radius, which is set to $0.2$ throughout the experiments.
The ego agent may choose to navigate to the watch tower for full observability.
During its contact with any of the watch towers, the ego agent can observe all the agents and landmarks in the environment.

\textbf{Observation space.} The observation space is a $37$-dimensional vector, consisting of the positions and velocities of the agents and the positions of the landmarks.
An additional 0/1 sign is added for every landmark and every agent other than the ego agent to indicate if the entity is currently visible by the ego agent.
Invisible entities have the sign, positions, and velocities set to $0$.
All positions, excluding that of the ego agent, are relative to the ego agent.

\textbf{Action space.} We use the discrete action space of MPE, with 5 actions corresponding to moving left/right/up/down and standing still.

\textbf{Reward.} The predators share a common reward that encourages them to collaborate and cover all the prey.
Specifically, denote $A$ as the set of all predators and $B$ as the set of all prey, the reward for predators at each time step is given as
$$
-c \sum_{b \in B} \min_{a \in A} d(a, b)
$$
where $d$ is the Euclidean distance function, $c=0.1$.
Intuitively, this reward allows the predators to divide and conquer such that for every prey there is a predator nearby.

\section{Peer Pool Generation} \label{sec:peer_pool_app}
In the PACE pipeline, we need to first collect a diverse peer pool $\boldsymbol{\Psi}$, which contains representative behaviors of the real peer distribution. 
Current methods~\citep{strouse2021collaborating, charakorn2023generating, lupu2021trajectory} mainly use RL algorithms with diversity objectives to train policies that exhibit various behaviors. In this paper, however, we generate a collection of rule-based policies. This is because the P2 policy in Kuhn Poker can be parameterized by two probabilities ($\xi,\eta$). We believe the preference-based policies in PO-Overcooked and Predator-Prey-W capture more human-like behaviors within the game. The details of the rule-based policy pool are listed below.

\subsection{Kuhn Poker}
As we mentioned above, we eliminate the dominant strategies for P2. Therefore, P2 policy can be determined by two factors: $\eta$ and $\xi$. $\eta$ is the probability of betting with Queen after P1 bets. $\xi$ is the probability of betting with Jack after P1 passes.

In this way, we can easily generate as many P2 policies as we want by randomly sampling $\xi$ and $\eta$. In this paper, we sample 40 P2 policies for training and 10 P2 policies for testing.

\subsection{PO-Overcooked} \label{sec:peer_pool_ovk_app}
In this paper, we generate preference-based peer agents that possess individual preferences for specific recipes. For instance, each peer agent may have a preference for a recipe such as Tomato \& Onion Salad. These peer agents are consistently positioned on the right side of the kitchen and interact exclusively with ingredients and dishes that align with their preferred recipe. For instance, a Tomato \& Onion Salad peer agent focuses on sub-tasks related to handling Tomato and Onion ingredients (chopped or fresh) or delivering dishes that exclusively contain these two ingredients.

At each time step, the agent evaluates whether its current sub-task is completed or not. If the sub-task remains unfinished, the agent determines the shortest path to the target position and navigates accordingly. On the other hand, if the sub-task is completed, the agent samples a new sub-task from its preferred set of sub-tasks.

In addition, there are two parameters that control more fine-grained strategies. $P_{\text{nav}}$ is the probability of moving right/left instead of up/down when there are multiple shortest paths. $P_{\text{ act}}$ is the probability of choosing a random action instead of the optimal action for the current sub-task. For example, suppose the peer is trying to put the Tomato on the counter. With probability $P_{\text{act}}$, it randomly chooses an action from the action space. With probability $1-P_{\text{act}}$, it chooses the optimal action (navigate or interact). 

We believe such rule-based agents exhibit behaviors that are more human-like than self-play agents trained by RL algorithms. First, cognition studies~\citep{etel2019theory,sher2014children} suggest that humans indeed act based on intentions and desires. Furthermore, self-play agents often have arbitrary conventions~\citep{hu2020other}. In overcooked, such conventions may be putting/taking ingredients and plates at a certain counter and refusing to interact with objects at different locations. However, these self-play conventions rarely appear in human behaviors. As a result, a preference-based policy is a better choice.

The overcooked scenario in this paper consists of 9 recipes. There are also two parameters $P_{\text{nav}}$ and $P_{\text{act}}$ that control more fine-grained strategies. When generating a new peer policy, we first uniformly sample its preferred recipe from the 9 recipes, and then randomly sample $P_{\text{nav}}$ and $P_{\text{act}}$. The training peer pool contains 18 policies and the testing peer pool contains 9 policies.

\subsection{Predator-Prey-W}

For the predator peer, we design policies that have a preference towards a specific prey.
The predator peer will always chase the preferred prey under full observation.

For the prey peers, we construct $8$ different patterns (Figure \ref{fig:pp_env}, dotted lines, \textcircled{1}-\textcircled{8}), where each prey peer moves back and forth along a preferred path.
We divide the set of paths into a train set (blue dotted lines, \textcircled{1}-\textcircled{4}) and a test set (red dotted lines, \textcircled{5}-\textcircled{8}).
The final train peer pool is generated by sampling different combinations of 1 predator peer and 2 train prey peers, while the test peer pool samples combinations of 1 predator peer and 2 test prey peers.
As a result, during online adaptation, the policies of all prey peers are unseen to the ego agent.
We sample $16$ combinations for training and $24$ combinations for testing.

\section{Algorithm Details}

\subsection{Online Adaptation Details} \label{sec:online_adp_app}

\begin{algorithm}[tb]
\caption{Online Adaptation Procedure}\label{alg:adapt}
\begin{algorithmic}[1]
\REQUIRE Online peer $\boldsymbol{\psi}$, adaptation horizon $N_{\text{ctx}}$, parameters $\theta$
\OUTPUT Average episodic return
\STATE $C^1 \gets \emptyset, R \gets 0, t \gets 0$ \STATE Reset the environment and get $o_0^1$
\WHILE{$N_\text{eps}$ episodes not reached}
\STATE Step the environment with $a^1_t \sim \pi_\theta(a|o^1_t, \chi_\theta(C^1))$ and $\mathbf{a}^{-1}_t \sim \boldsymbol{\psi}$, obtain task reward $r_t$ and next observation $o^1_{t+1}$
\STATE Update $C^1$ with $(o^1_t, a^1_t)$
\STATE Update $R \gets R + r_t, t \gets t + 1$
\IF{Time to switch peer agents}
    \STATE Resample $\boldsymbol{\psi}$
\ENDIF
\IF{Criterion for clearing the context is met}
    \STATE $C^1 \gets \emptyset$
\ENDIF
\ENDWHILE
\STATE Return $\frac{R}{N_\text{eps}}$
\end{algorithmic}
\end{algorithm}

We present the pseudocode for the online adaptation procedure in Algorithm \ref{alg:adapt}.
The online adaptation procedure computes the total return of the task reward without the exploration reward (Line 6).
In the sudden-change peer experiment, the online peer may change (Line 8) and the context may be cleared (Line 11) during adaptation, but the two events take place separately, and the agent is unaware of the peer change except through the procedure described in Appendix \ref{sec:dyn_peer_app}.

\subsection{Training Details} \label{sec:train_details}

\begin{table}[t]
  \caption{Hyperparameters for all the algorithms in the Kuhn Poker environment.}
  \label{tab:hyp_kp}
  \centering
      \scalebox{1.0}{
  \begin{tabular}{cccccc}
    \toprule
    \multirow{2}{*}{Parameter Name} & \multicolumn{5}{c}{Algorithms}                   \\
    \cmidrule(r){2-6}
     & PACE & Generalist & LILI & LIAM(X) &  GSCU \\
    \midrule
    Learning Rate & 2e-4 & 2e-4 & 2e-4 & 2e-4 & 5e-4 \\
    PPO Clip $\epsilon$ & $0.2$ & $0.2$ & $0.2$ & $0.2$ & $0.2$ \\
    Entropy Coefficient & 5e-4 & 5e-4 & 5e-4 & 5e-4 & $0.01$ \\
    $\gamma$ & $0.99$ & $0.99$ & $0.99$ & $0.99$ & $0.99$ \\
    GAE $\lambda$ & $0.95$ & $0.95$ & $0.95$ & $0.95$ & $0.95$ \\
    Batch Size & $80000$ & $80000$ & $80000$ & $80000$ & $1000$ \\
    \# Update Epochs & $15$ & $15$ & $15$ & $15$ & $5$ \\
    \# Mini Batches & $12$ & $12$ & $12$ & $12$ & $30$ \\
    Gradient Clipping (L2) & $2.0$ & $2.0$ & $2.0$ & $2.0$ & $0.5$ \\
    Activation Function & ReLU & ReLU & ReLU & ReLU & ReLU \\
    Actor/Critic Hidden Dims & [128, 128] & [128, 128] & [128, 128] & [128, 128] & [128, 128] \\
    $f_\theta$ Hidden Dims & [64, 64] & N/A & N/A & N/A & N/A \\
    $g_\theta$ Hidden Dims & [64] & N/A & N/A & N/A & N/A \\
    \bottomrule
  \end{tabular}
  }
\end{table}

\begin{table}[t]
  \caption{Hyperparameters for all the algorithms in the PO-Overcooked environment.}
  \label{tab:hyp_ovk}
  \centering
  \scalebox{1.0}{
  \begin{tabular}{cccccc}
    \toprule
    \multirow{2}{*}{Parameter Name} & \multicolumn{5}{c}{Algorithms}                   \\
    \cmidrule(r){2-6}
     & PACE     & Generalist & LILI & LIAM(X) &  GSCU \\
    \midrule
    Learning Rate & 1e-3 & 1e-3 & 1e-3 & 1e-3 & 7e-4 \\
    PPO Clip $\epsilon$ & $0.2$ & $0.2$ & $0.2$ & $0.2$ & $0.2$ \\
    Entropy Coefficient & $0.03$ & $0.03$ & $0.03$ & $0.03$ & $0.01$ \\
    $\gamma$ & $0.99$ & $0.99$ & $0.99$ & $0.99$ & $0.99$ \\
    GAE $\lambda$ & $0.95$ & $0.95$ & $0.95$ & $0.95$ & $0.95$ \\
    Batch Size & $72000$ & $72000$ & $72000$ & $72000$ & $2500$ \\
    \# Update Epochs & $4$ & $4$ & $4$ & $4$ & $8$ \\
    \# Mini Batches & $18$ & $18$ & $18$ & $18$ & $2$ \\
    Gradient Clipping (L2) & $15.0$ & $15.0$ & $15.0$ & $15.0$ & $0.5$ \\
    Activation Function & ReLU & ReLU & ReLU & ReLU & Tanh \\
    Actor/Critic Hidden Dims & [128, 128] & [128, 128] & [128, 128] & [128, 128] & [64 64] \\
    $f_\theta$ Hidden Dims & [128, 128] & N/A & N/A & N/A & N/A \\
    $g_\theta$ Hidden Dims & [128] & N/A & N/A & N/A & N/A \\
    \bottomrule
  \end{tabular}
  }
    \end{table}

\begin{table}[t]
  \caption{Hyperparameters for all the algorithms in the Predator-Prey-W environment.}
  \label{tab:hyp_pp}
  \centering
  \scalebox{1.0}{
  \begin{tabular}{cccccc}
    \toprule
    \multirow{2}{*}{Parameter Name} & \multicolumn{5}{c}{Algorithms}                   \\
    \cmidrule(r){2-6}
     & PACE     & Generalist & LILI & LIAM(X) &  GSCU \\
    \midrule
    Learning Rate & 1e-3 & 1e-3 & 1e-3 & 1e-3 & 5e-4 \\
    PPO Clip $\epsilon$ & $0.2$ & $0.2$ & $0.2$ & $0.2$ & $0.2$ \\
    Entropy Coefficient & $0.03$ & $0.03$ & $0.03$ & $0.03$ & $0.01$ \\
    $\gamma$ & $0.99$ & $0.99$ & $0.99$ & $0.99$ & $0.99$ \\
    GAE $\lambda$ & $0.95$ & $0.95$ & $0.95$ & $0.95$ & $0.95$ \\
    Batch Size & $64000$ & $64000$ & $64000$ & $64000$ & $2500$ \\
    \# Update Epochs & $4$ & $4$ & $4$ & $4$ & $8$ \\
    \# Mini Batches & $16$ & $16$ & $16$ & $16$ & $2$ \\
    Gradient Clipping (L2) & $15.0$ & $15.0$ & $15.0$ & $15.0$ & $0.5$ \\
    Activation Function & ReLU & ReLU & ReLU & ReLU & Tanh \\
    Actor/Critic Hidden Dims & [128, 128] & [128, 128] & [128, 128] & [128, 128] & [64 64] \\
    $f_\theta$ Hidden Dims & [128, 128] & N/A & N/A & N/A & N/A \\
    $g_\theta$ Hidden Dims & [128] & N/A & N/A & N/A & N/A \\
    \bottomrule
  \end{tabular}
  }
    \end{table}

The general training pipeline of LILI, LIAM, LIAMX, and Generalist is similar to PACE in Algorithm~\ref{alg:train}. The difference between these methods and PACE is that they have some different auxiliary tasks and do not have the exploration reward used in PACE. The training procedure of GSCU is quite different from other methods, which can be found in the original paper.

For all baselines and ablations, we use PPO~\citep{schulman2017proximal, pytorchrl} as the RL training algorithm.
Table \ref{tab:hyp_kp}, \ref{tab:hyp_ovk}, and \ref{tab:hyp_pp} list the hyperparameters related to architectures and PPO training for Kuhn Poker, PO-Overcooked, and Predator-Prey-W, respectively.

We keep the original hyperparameters for GSCU on Kuhn Poker.
For Kuhn Poker, the training budget for all algorithms except GSCU is 5 million steps, while for GSCU embedding learning takes 1 million episodes and conditional RL takes 1 million episodes.
For PO-Overcooked, the training budget for all algorithms except GSCU is 30 million steps, while for GSCU the embedding learning takes 2 million steps and the conditional RL takes 30 million steps.
For Predator-Prey-W, the training budget for all algorithms including GSCU is 15 million steps.
The embedding learning for GSCU takes an additional 2 million steps.
The training of PACE takes $\sim 12$ hours with $\sim 80$ processes on a single Titan Xp GPU.
Both the PACE actor $\pi_\theta(a|o, \chi_\theta(C))$ and critic take concatenated observation and encoder output as the input.

For algorithms using RNN, including Generalist, LIAM, and LIAMX, the RNN is implemented as a single-layer GRU with $128$ hidden units.
The RNN is trained using back-propagation through time (BPTT) with gradients detached every $20$ steps.
Actor and critic share the same RNN, as well as the hidden layers before the RNN.

For methods with auxiliary tasks, there is an additional loss accompanying the main RL loss, computed using the same mini-batch as used in RL training.
For PACE, the auxiliary loss is used with a weight $1.0$.
For LIAM, the auxiliary loss is used with weight $1.0$ for both action and observation prediction.
For LILI, the context is the last episode, as specified in~\cite{xie2021learning}.
The auxiliary loss is used with weight $1.0$ for both reward and next observation prediction.

The coefficient for PACE's exploration reward decays from $c_\text{init}$ to $0$ in $M$ steps.
$c_\text{init}=0.2$ for PO-Overcooked, $0.01$ for Kuhn Poker, and $0.1$ for Predator-Prey-W, while $M=2.5*10^7$ for PO-Overcooked, $4*10^6$ for Kuhn Poker, and $1.5*10^7$ for Predator-Prey-W.
We choose these hyperparameters such that the exploration reward has an initial scale similar to that of the task reward, and decays to $0$ near the end of the training.
Additionally, we warm up the context encoder for $M_w$ steps using the auxiliary loss only without RL loss.
$M_w=10^6$ for PO-Overcooked, $10^5$ for Kuhn Poker, and $5*10^5$ for Predator-Prey-W.

\subsection{Adapting to the Sudden-change Peer } \label{sec:dyn_peer_app}

\begin{figure}[t]
    \centering
    \includegraphics[width=0.5\textwidth]{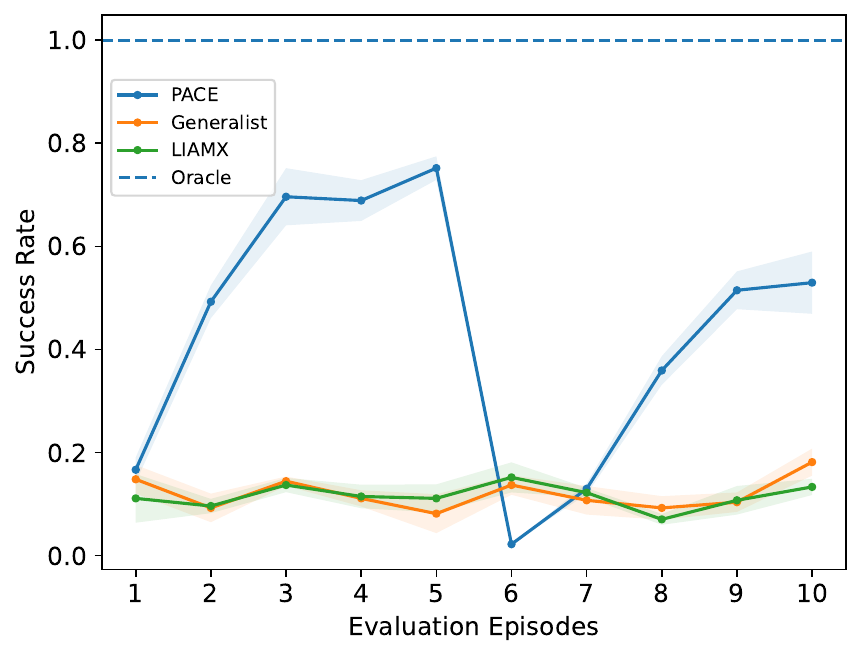}
    \caption{Results of peers with sudden changes in PO-Overcooked. After the peer change in episode 6, PACE detects the change and clears the context, allowing performance to recover, while baselines fail consistently.}
    \label{fig:dyn_peer}
\end{figure}

We present more details of the sudden-change peer experiment in section \ref{sec:dyn_peer}.
We run the experiment in PO-Overcooked for $10$ episodes.
The ego agent needs to collaborate with two different peers in the first $5$ episodes and the last $5$ episodes without knowing when the peer change takes place.
While the PACE agents are trained against static peers that never change, we use drops in the evaluation metric as an indicator of peer changes and clear the context when such changes take place.
Formally, denote the evaluation metric for episode $i$ as $R_i$, then we detect a peer change at $i$ iff

$$
R_i < c_{th} * \max_{j \leq i} R_j  + (1 - c_{th}) * \min_{j \leq i} R_j
$$

where $c_{th} \in [0, 1]$ is a threshold coefficient.
$c_{th}=1$ is the most aggressive (clears context whenever performance drops) and $c_{th}=0$ is the most conservative (never clears the context).
We set $c_th = 0.8$ in this experiment.

The success rate curve during the adaptation procedure is presented in Figure \ref{fig:dyn_peer}.
PACE successfully recovers after the change of peer agent in episode 6, while baselines consistently underperform.

\section{Additional Experiments}

\begin{figure*}[t]
    \centering
    \begin{subfigure}{0.45\textwidth}
        \includegraphics[width=\linewidth]{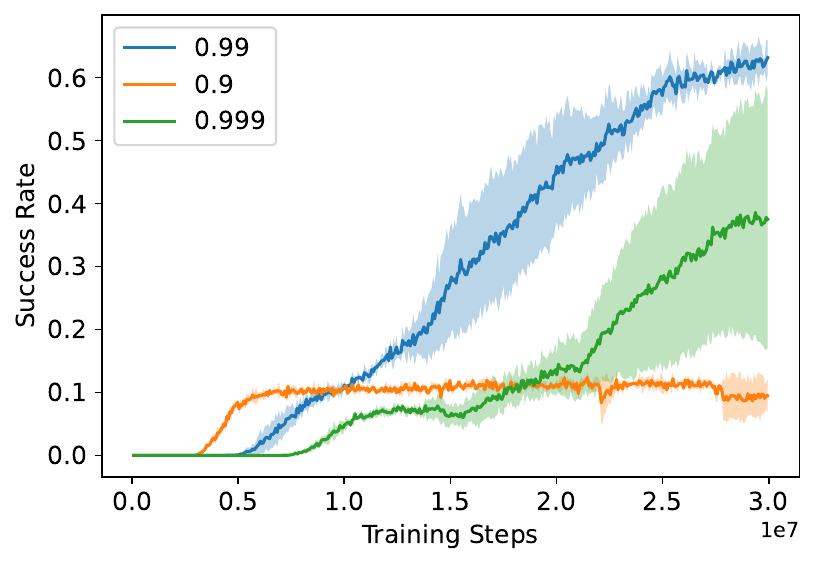}
        \caption{Success Rate}
        \label{fig:gamma_sr}
    \end{subfigure}
                    \begin{subfigure}{0.45\textwidth}
        \includegraphics[width=\linewidth]{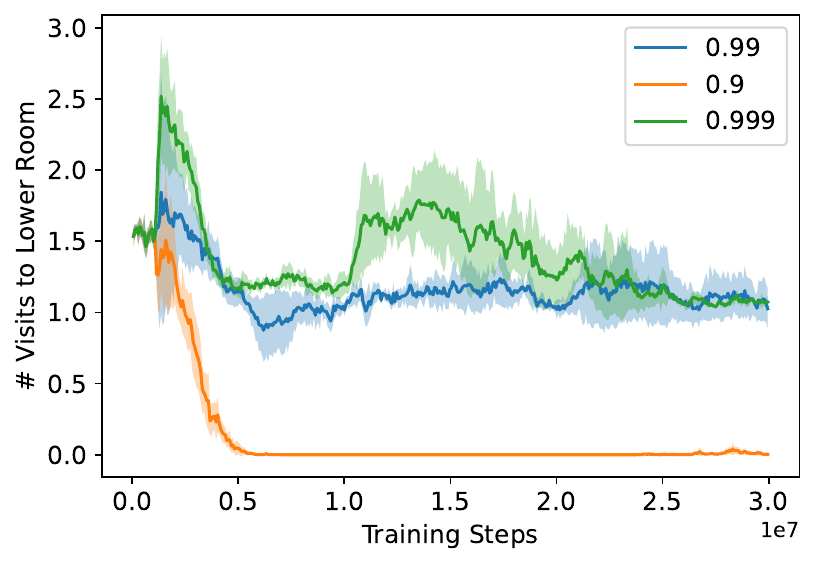}
        \caption{Number of Visits to the Lower Room}
        \label{fig:gamma_visit}
    \end{subfigure}
 	                  \caption{Training success rates (a) and number of visits to the lower room (b) for the discount factor ablations in PO-Overcooked. Our discount factor of choice, $\gamma=0.99$, performs best. $\gamma=0.9$ is short-sighted and does not exhibit exploration behaviors, while $\gamma=0.999$ introduces training instabilities.}
    \label{fig:gamma}
\end{figure*}

\subsection{Impact of Discount Factor}

\begin{figure*}[t]
    \centering
    \begin{subfigure}{0.45\textwidth}
        \includegraphics[width=\linewidth]{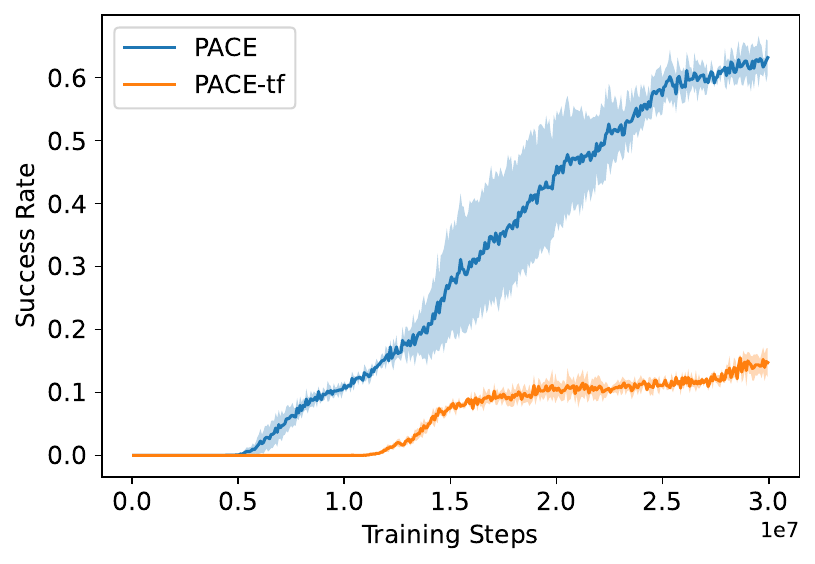}
        \caption{Success Rate}
        \label{fig:tf_sr}
    \end{subfigure}
                    \begin{subfigure}{0.45\textwidth}
        \includegraphics[width=\linewidth]{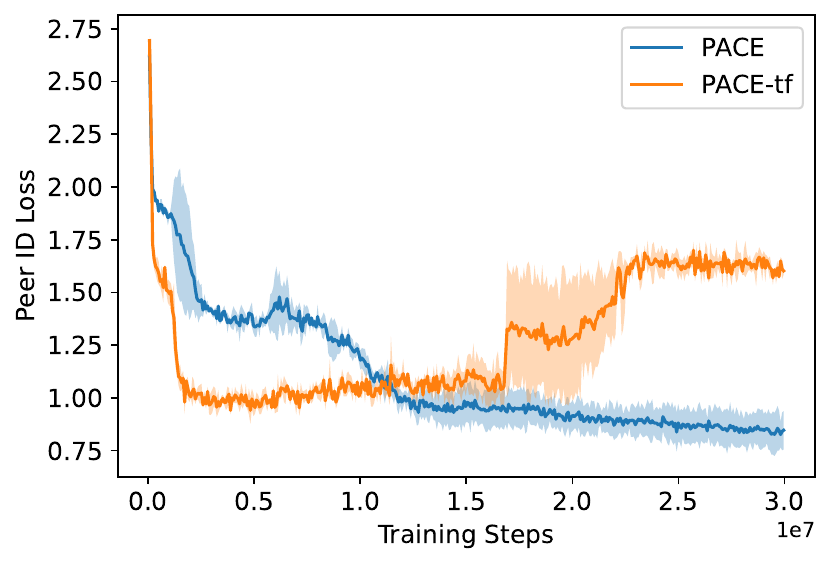}
        \caption{Peer Identification Loss}
        \label{fig:tf_loss}
    \end{subfigure}
 	                  \caption{Training success rates (a) and peer identification loss (b) for the Transformer encoder ablations in PO-Overcooked. PACE-tf is }
    \label{fig:tf}
\end{figure*}

We conduct an ablation in PO-Overcooked for the impact of the discount factor.
The discount factor $\gamma$ defines the RL objective \ref{eq:meta_return}, where a small $\gamma$ puts more emphasis on short-term returns and a large $\gamma$ approximates the undiscounted return better but has higher variance.
Figure \ref{fig:gamma_sr} and \ref{fig:gamma_visit} present the training success rates and number of visits to the lower room.
The number of visits to the lower room reflects the exploratory tendencies of the agent since this action only reveals information (behaviors of the peer agent in the lower room) to the agent instead of generating imminent environment rewards. 
As seen in Figure \ref{fig:gamma}, $\gamma=0.9$ leads to a short-sighted policy that quickly converges to a local minimum, even in the presence of an exploration reward.
A larger $\gamma=0.999$ retains the exploration behavior but leads to unstable training.

\subsection{Impact of Transformer Encoder}

We designed a bi-level transformer encoder to compare with the MLP encoder we used in our primary experiments.
We construct two transformer encoders for encoding episode-level information and context-level information, respectively.
Formally, the encoding for episode $n$ is

$$
z^1_n := Enc^e((o_{n, 1}^1, a_{n, 1}^1), (o_{n, 2}^1, a_{n, 2}^1), \ldots, (o_{n, T_n}^1, a_{n, T_n}^1))
$$

while the encoding for the whole context $C$ is 

$$
z^1 := Enc^c(z^1_1, z^1_2, \ldots, z^1_N)
$$

where $Enc^e, Enc^c$ are transformer encoders for the episode- and context-level, followed by average pooling over the sequence dimension.
We apply learnable positional embedding and standard regularization techniques like Dropout to the encoder.
We use hidden dimension and feed-forward dimension of $128$, Dropout coefficient of $0.5$, $4$ heads for the multi-head attention, and a single layer of transformer encoder in each of $Enc^e$ and $Enc^c$.

As shown in Figure~\ref{fig:tf}, PACE-tf is greatly outperformed by the standard PACE MLP encoder.
The reason for this underperformance, we hypothesize, is that the transformer encoder may overfit the current RL training batch on the peer identification task.
In Figure~\ref{fig:tf_loss}, it can be seen that the peer ID loss of PACE-tf drops faster than PACE initially, but slowly increases after that, until an abrupt change in the middle of the training.
In contrast, the loss for PACE drops consistently along the training procedure.
It is worth noting that our peer identification task and exploration reward are not limited to a specific kind of encoder architecture, and that we choose MLP with average pooling based on empirical performance.

\subsection{Impact of Training Horizon}
\label{sec:abl_horizon}

The goal of peer adaptation is to optimize the total return over $N_{eps}$ episodes of interaction with an unknown peer.
Different $N_{eps}$ may induce different problem instances, and correspondingly, different optimal strategies.
For example, when $N_{eps}$ is small, the agent may not have much time to explore and instead settle for an exploitative strategy that doesn't reveal interesting contexts.
In contrast, when $N_{eps}$ is large, the agent will have a chance to select exploratory actions that are suboptimal in terms of instant rewards but make up for the shortfall by guiding many subsequent episodes of exploitation.

To demonstrate the impact of the horizon, we adjust the context length and return computation horizon during training (termed $N_{ctx}$ below), which could potentially be different from the testing horizon (termed $N_{eps}$, as in the main text).
We conduct an experiment in Kuhn Poker to train PACE agents with $N_{ctx} \in \{20, 50, 100, 150\}$ and cross-test them with a new $N_{eps}$, where $N_{ctx}=N_{eps}=100$ is our original setting.

\begin{table}[!t]
    \caption{Average return over $N_{eps}$ testing episodes of PACE agents trained with $N_{ctx}$ episodes of context in Kuhn Poker.}
    \label{tab:abl_horizon}
    \centering
    \begin{tabular}{ccccc}
    \toprule
        ~ & $N_{ctx}=20$ & $N_{ctx}=50$ & $N_{ctx}=100$ & $N_{ctx}=150$ \\ \midrule
        $N_{eps}=20$ & \textbf{0.029 $\pm$ 0.001} & 0.024 $\pm$ 0.003 & 0.022 $\pm$ 0.003 & 0.018 $\pm$ 0.004 \\         $N_{eps}=50$ & 0.039 $\pm$ 0.001 & \textbf{0.040 $\pm$ 0.003} & \textbf{0.040 $\pm$ 0.004} & 0.038 $\pm$ 0.002 \\         $N_{eps}=100$ & 0.044 $\pm$ 0.003 & 0.047 $\pm$ 0.003 & \textbf{0.049 $\pm$ 0.003} & 0.048 $\pm$ 0.002 \\         $N_{eps}=150$ & 0.046 $\pm$ 0.004 & 0.050 $\pm$ 0.002 & \textbf{0.053 $\pm$ 0.004} & \textbf{0.053 $\pm$ 0.002} \\ \bottomrule
    \end{tabular}
    
\end{table}

The results are presented in Table \ref{tab:abl_horizon}.
It can be seen that the agent trained with a short horizon learns to exploit early in the game at the cost of a diminished future return due to the lack of information (e.g. $N_{ctx}=20$ achieves the highest average return of 0.029 in the first 20 episodes, but in subsequent episodes fails to exploit as well as agents with a larger $N_{ctx}$), while an agent trained with a long horizon sacrifices returns of early episodes to improve overall performance (e.g. $N_{ctx}=150$ is bad in the first 20 episodes but good over all 150 episodes).
For each $N_{eps}$, the highest average return over that horizon is always achieved by the agent trained with $N_{ctx}=N_{eps}$, indicated by the bold numbers across the diagonal of the table.

\subsection{Peer Scalability}

\begin{table}[!t]
    \centering
    \caption{Average returns of PACE and LIAMX agents tested with 3 peers and trained under different settings in Predator-Prey-W.}
    \label{tab:transferpeers}
    \begin{tabular}{ccc}
    \toprule
        Testing Setting & PACE & LIAMX \\ \midrule
        Train $m=3$ (In-domain) & -3.93 $\pm$ 0.04 & -4.34 $\pm$ 0.08 \\         Train $m=5$ (Transfer) & -4.08 $\pm$ 0.08 & -4.63 $\pm$ 0.04 \\ \bottomrule
    \end{tabular}
\end{table}

In this section, we conduct further experiments to validate PACE's scalability with respect to the number of peer agents.
PACE can efficiently handle different numbers of peer agents (denoted by $m$), ranging from $m=1$ (Kuhn Poker with one opponent and PO-Overcooked with one teammate) to $m=3$ (Predator-Prey-W with one predator and two preys) in our main experiments.
Here we scale the number of peer agents to 5 in Predator-Prey-W, resulting in 2 peer predators and 3 prey.
The horizon $N_{eps}$ is kept unchanged at 5 episodes.
In this scenario, our PACE agent achieves an average reward of $-4.346 \pm 0.045$, while our strongest baseline LIAMX obtains $-4.820 \pm 0.016$, a gap of $0.474$ larger than that in the main experiment ($0.410$).

Furthermore, we additionally investigate the transfer setting, where the number of peer agents differs between training and testing.
Changing the number of peers between training and testing may dramatically alter the dynamics of a game, making the adaptation problem even harder.
For example, the dimension of the observation may change with the number of peers, so an architecture that can handle a variable number of peer inputs is required.
Moreover, we may need to randomize the number of peers during training to increase the game dynamic diversity, which helps mitigate the gap between training and evaluation.
While we plan to explore this more thoroughly in future work, we conduct a preliminary experiment in Predator-Prey-W by testing agents trained under $m=3,5$ (described above) in a $m=3$ scenario (same as the main experiment).
When testing in the transfer setting, the discrepancy in observation size is mitigated by padding the observation with two dummy invisible agents.
As shown in Table~\ref{tab:transferpeers}, the PACE agent trained with an in-domain $m=3$ performs better than the transferred PACE agent trained under $m=5$, but the gap is small (-3.93 versus -4.08).
In both the in-domain and the transfer scenario, PACE agents outperform the LIAMX agents.
These results validate the scalability of PACE agents with respect to the number of peer agents.

\subsection{Pool Diversity}

\begin{table}[!t]
    \caption{Average success rates of PACE agents trained with different numbers of recipes in the training peer pool.}
    \label{tab:pool_diversity}
    \centering
    \begin{tabular}{ccccc}
    \toprule
        ~ & 2 recipes & 4 recipes & 6 recipes & 9 recipes \\ \midrule
        Success Rate & 0.193 $\pm$ 0.012 & 0.392 $\pm$ 0.019 & 0.486 $\pm$ 0.017 & 0.553 $\pm$ 0.029 \\ \bottomrule
    \end{tabular}
\end{table}

The diversity of the peer pool is critical for the performance of peer adaptation algorithms. 
To demonstrate this, in PO-Overcooked, we control the diversity of the training pool by adjusting the number of recipes (originally 9 recipes) used in training.
Specifically, we reduce the number of distinct recipes in the training peer pool while retaining its size (18).
The testing pool is left unchanged. 
The results are shown in Table~\ref{tab:pool_diversity}. 
It can be seen that while all agents are trained with 18 peers, the decrease in diversity causes the performance to drop, and the magnitude of the drop is correlated with the pool diversity. 
This indicates that our method can benefit from the increasing diversity of the training pool.

\subsection{Preliminary Human Study}

Developing agents capable of helpful and safe interactions with humans in real-world scenarios is a longstanding goal of the multi-agent field.
In PACE, we strive to approximate sophisticated real-world scenarios with partial observability, one of the common complexities of the real world.
We also designed rule-based peer policies that emulate diverse real-world human behaviors.
Furthermore, we conducted a preliminary human study to demonstrate the capability of PACE agents to adapt to and cooperate with human players.
We pair PACE and LIAMX agents with 10 human participants to collaborate in the PO-Overcooked environment for a total of 54 trials with 5 episodes each.
As a result, the average success rate for PACE agent is $0.504 \pm 0.046$, while the average success rate for LIAMX agent is $0.074 \pm 0.028$.
The PACE agent maintains a decent success rate that is broadly in line with the results of the main experiment, outperforming the LIAMX baseline.
This validates the ability of PACE agents to adapt to human behaviors and strategies.

\end{document}